\documentclass[11pt,letterpaper]{article}
\usepackage[margin=1in]{geometry}
\usepackage{times,amsmath,amssymb,booktabs,graphicx}
\usepackage[dvipsnames]{xcolor}
\usepackage[numbers,sort&compress]{natbib}
\usepackage[font=small,labelfont=bf]{caption}
\usepackage[inline,shortlabels]{enumitem}
\usepackage{microtype}
\usepackage{tabularx}
\usepackage{placeins}
\usepackage{tikz}
\usetikzlibrary{arrows.meta,positioning,fit,backgrounds,calc}

\newcommand{\pizero}{\ensuremath{\pi_{0.5}}}
\newcommand{\rmr}{RMR-VLA}
\newcommand{\best}[1]{\textbf{#1}}
\newcommand{\second}[1]{\underline{#1}}
\newcolumntype{Y}{>{\raggedright\arraybackslash}X}

\definecolor{rmrblue}{HTML}{285F8F}
\definecolor{rmrteal}{HTML}{237A70}
\definecolor{rmrpurple}{HTML}{6C5593}
\definecolor{rmrgray}{HTML}{45505A}
\definecolor{rmrlight}{HTML}{F0F5F9}
\definecolor{rmrevidence}{HTML}{256BA2}
\definecolor{rmrevent}{HTML}{B97A20}
\definecolor{rmrstate}{HTML}{7753AE}
\definecolor{rmrgoal}{HTML}{25846B}

\usepackage{xurl}
\usepackage[breaklinks,colorlinks,allcolors=rmrblue]{hyperref}
\usepackage[capitalize]{cleveref}
\crefname{section}{Sec.}{Secs.}
\Crefname{section}{Section}{Sections}
\crefname{table}{Table}{Tables}
\setlist[itemize]{noitemsep,leftmargin=*,topsep=0pt}

\title{RMR-VLA: Recurrent Memory Reasoning for Vision-Language-Action Models}
\author{Yiheng Du\textsuperscript{1,*}\quad Jingqi Tian\textsuperscript{2,*}\quad
    Xu Jiang\textsuperscript{1,*}\quad Jiexuan Zhang\textsuperscript{1}\\[3pt]
    Shuyu Wang\textsuperscript{1}\quad Zecheng Tang\textsuperscript{1}\quad Jian Zhang\textsuperscript{1}}
\makeatletter
\providecommand\@LN[2]{}
\hypersetup{pdftitle={\@title},
    pdfauthor={Yiheng Du, Jingqi Tian, Xu Jiang, Jiexuan Zhang, Shuyu Wang, Zecheng Tang, Jian Zhang}}
\renewcommand{\maketitle}{%
    \begin{center}
    {\LARGE\bfseries \@title\par}
    \vspace{1.2em}
    {\large \@author\par}
    \vspace{0.6em}
    \textsuperscript{1}Peking University\qquad\textsuperscript{2}Tsinghua University
    \end{center}
    \vspace{0.6em}}
\makeatother

\begin{document}
\maketitle
\begingroup
\renewcommand{\thefootnote}{\fnsymbol{footnote}}
\footnotetext[1]{These authors contributed equally to this work.}
\endgroup
\begin{abstract}
Long-horizon, history-dependent robotic manipulation often requires acting on visual cues that have disappeared, such as an object indicated earlier or a target hidden under a moving cup.
The key challenge is to maintain what past evidence means for the next action as the scene changes.
We introduce \emph{Recurrent Memory Reasoning} (RMR), an explicit reasoning step that connects memory to action inside a vision-language-action model (VLA), with no external reasoning model.
At each update, the VLA first infers an event from visual evidence, then revises the task state and grounds a subgoal in the current scene to guide action generation.
Carrying the state and subgoal across updates lets remembered identities and task progress determine the current target, doubling RoboMME success over an otherwise identical model without recurrence.
Joint training lets one VLA both reason and act; because its memory decisions are text tokens with exact likelihoods, GRPO refines them from rollout rewards without making action generation stochastic.
Among the compared methods, \rmr{} achieves the best results on RoboMME (66.46\% overall, 79.83\% on the Reference category) and LIBERO-Mem (90.08\%).
\end{abstract}

\begin{figure}[!h]
    \centering
    \includegraphics[width=\textwidth]{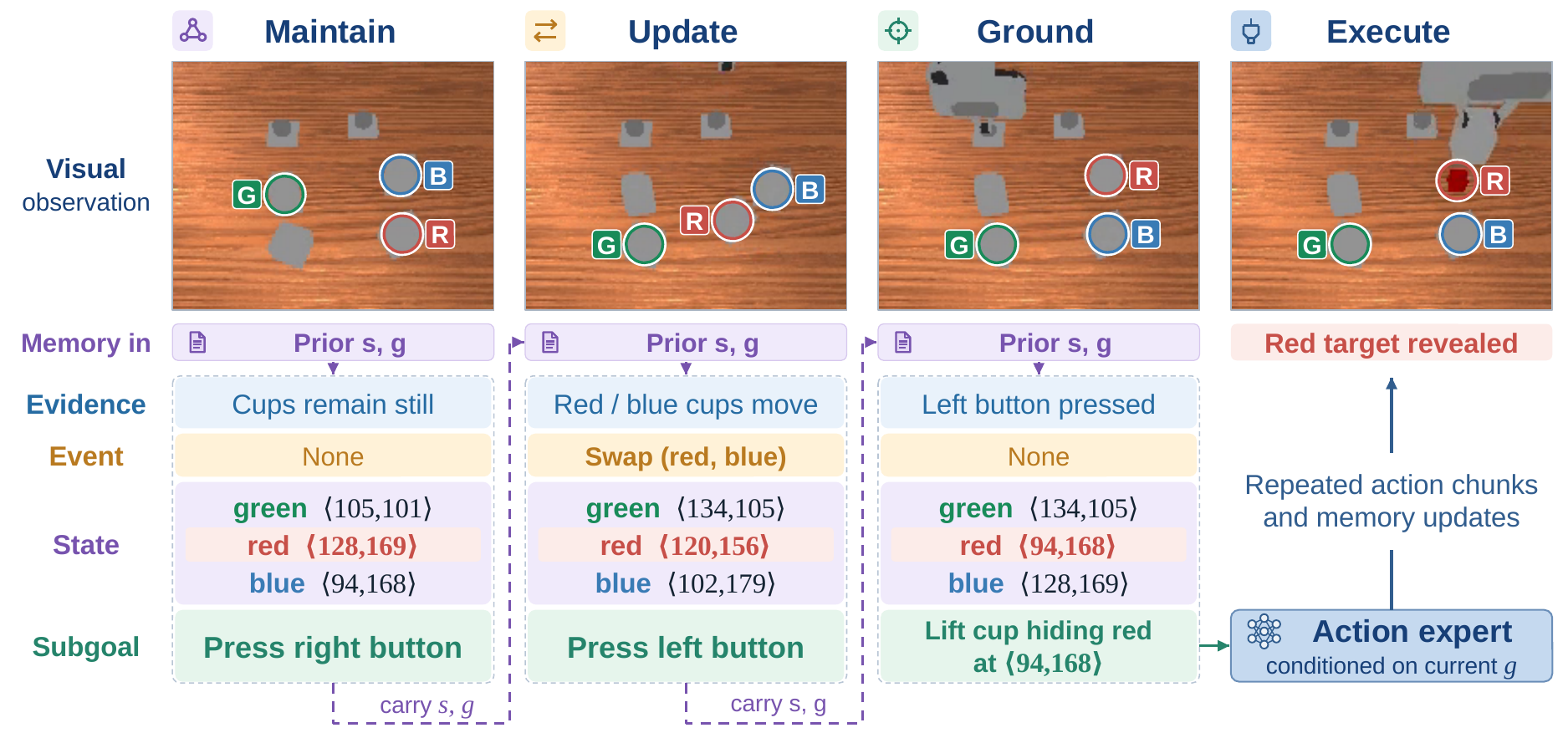}
    \caption{\textbf{Tracking hidden targets guides action.}
    In this example from RoboMME's ButtonUnmaskSwap task~\cite{robomme}, red's updated location grounds the lifting subgoal after a cup swap.
    G/R/B markers follow cube identities; evidence, event, and subgoal are abbreviated from the model's outputs.
    State tracks cube identities with image coordinates.
    Dashed links span intervening updates; red is revealed after repeated action chunks and memory updates.
    Coordinates $\langle y,x\rangle$ index rows and columns of the original $256\!\times\!256$ image.}
    \label{fig:teaser}
\end{figure}
\section{Introduction}
\label{sec:intro}

The visual evidence that determines a robot's next action can disappear before the robot is ready to act.
Consider a red cube hidden under a cup that then trades places with another cup.
Choosing which cup to lift requires linking red to the cup that now hides it.
Similar demands arise when an indicated object loses its highlight or when a repeated manipulation returns the scene to a familiar configuration.
In such long-horizon, history-dependent robotic manipulation~\cite{robomme,liberomem}, the next action depends on task information that the current image does not reveal.

Memory-augmented vision-language-action models (VLAs) retain temporal context through sampled or compressed visual representations or through language records~\cite{robomme,divide,dualstream,ltm,t2mem,notestoself,tacitvla,unimem}, while agentic systems reason over history with external models~\cite{ponderpounce,gamma,pal}.
When visual histories or latent memories feed action prediction directly, the policy must infer from the accumulated context how object associations and task progress have changed.
Language records make spatial context, plans, and attempted or completed subtasks explicit~\cite{notestoself,tacitvla}, yet the policy must still resolve these entries against the changed scene.
Both leave the policy to work out, while acting, what past information implies now; agentic systems reason about it explicitly, but in an external model.

What must be carried across decisions is twofold: the task state, since an object association can stay relevant across many actions, and the operation under way, since its completion advances progress and changes the next target.

We introduce \emph{Recurrent Memory Reasoning} (RMR), which turns each memory update into explicit reasoning inside the VLA.
At each update, its vision-language model (VLM) writes one reasoning trace: from observed \emph{evidence} it infers an \emph{event}, revises the task \emph{state}, and grounds the next \emph{subgoal} in it.
The state and subgoal feed the next update, and the subgoal guides the next actions.
In \cref{fig:teaser}, the cups hiding red and blue swap while the robot presses a button; the trace records the swap and red's new location, and once the button is pressed, the VLM reasons that the next subgoal is to lift the cup now hiding red.
Each update thus reasons out what past information implies now, before the robot acts.

RMR reasons within the VLA, with no external reasoning model.
We build \rmr{} on \pizero~\cite{pi05}, whose VLM already predicts a semantic subtask that conditions its action expert; in \rmr{}, this subtask becomes the subgoal reasoned out by the memory update.
The central difficulty is that the VLM must serve two roles, writing the reasoning trace and encoding inputs for the action expert, and training them one after the other weakens one of them.
We therefore train trace generation and action prediction jointly, then freeze the VLM and fine-tune the action expert.

Since the deployed model reads the state it wrote itself, we refine the model by reinforcement learning (RL) on its own rollouts.
Flow-based VLAs such as \pizero{} give no tractable action likelihood, so $\pi$RL recasts their action generation as a stochastic process before applying RL~\cite{pirl}.
\rmr{} makes its memory decisions in text with exact token likelihoods, so group relative policy optimization (GRPO)~\cite{shao2024deepseekmath,guo2025deepseekr1} optimizes the reasoning trace directly while the action expert stays frozen.

\rmr{} is strongest where earlier observations determine later targets, setting the best results among the compared methods on the RoboMME Reference tasks~\cite{robomme} (79.83\%) and on LIBERO-Mem~\cite{liberomem} (90.08\%).
Across all RoboMME tasks, its 66.46\% is the best result among the compared methods, including systems with an external reasoning model.
Carrying the state and subgoal forward doubles success over an otherwise identical model without recurrence, and among the trace fields the maintained state brings the largest gain, doubling the correctness of subgoals that act on earlier cues.

Our contributions are threefold:
\begin{itemize}[leftmargin=*,nosep]
    \item \textbf{Recurrent memory reasoning.} Each memory update is an explicit reasoning chain from visual evidence through events and a maintained task state to a grounded subgoal, revising what changes and preserving what still holds.
    \item \textbf{Memory reasoning and control in one VLA.} The VLA's own VLM reasons out the subgoal for its action expert; joint training and action-expert fine-tuning preserve both roles, and GRPO optimizes the text trace directly.
    \item \textbf{Performance and mechanism validation.} The best RoboMME and LIBERO-Mem results among the compared methods, including systems with an external reasoning model, with controlled comparisons of recurrence, trace structure, training, and reward.
\end{itemize}

\section{Related Work}
\label{sec:related}

\begin{figure*}[t]
    \centering
    \includegraphics[width=\textwidth]{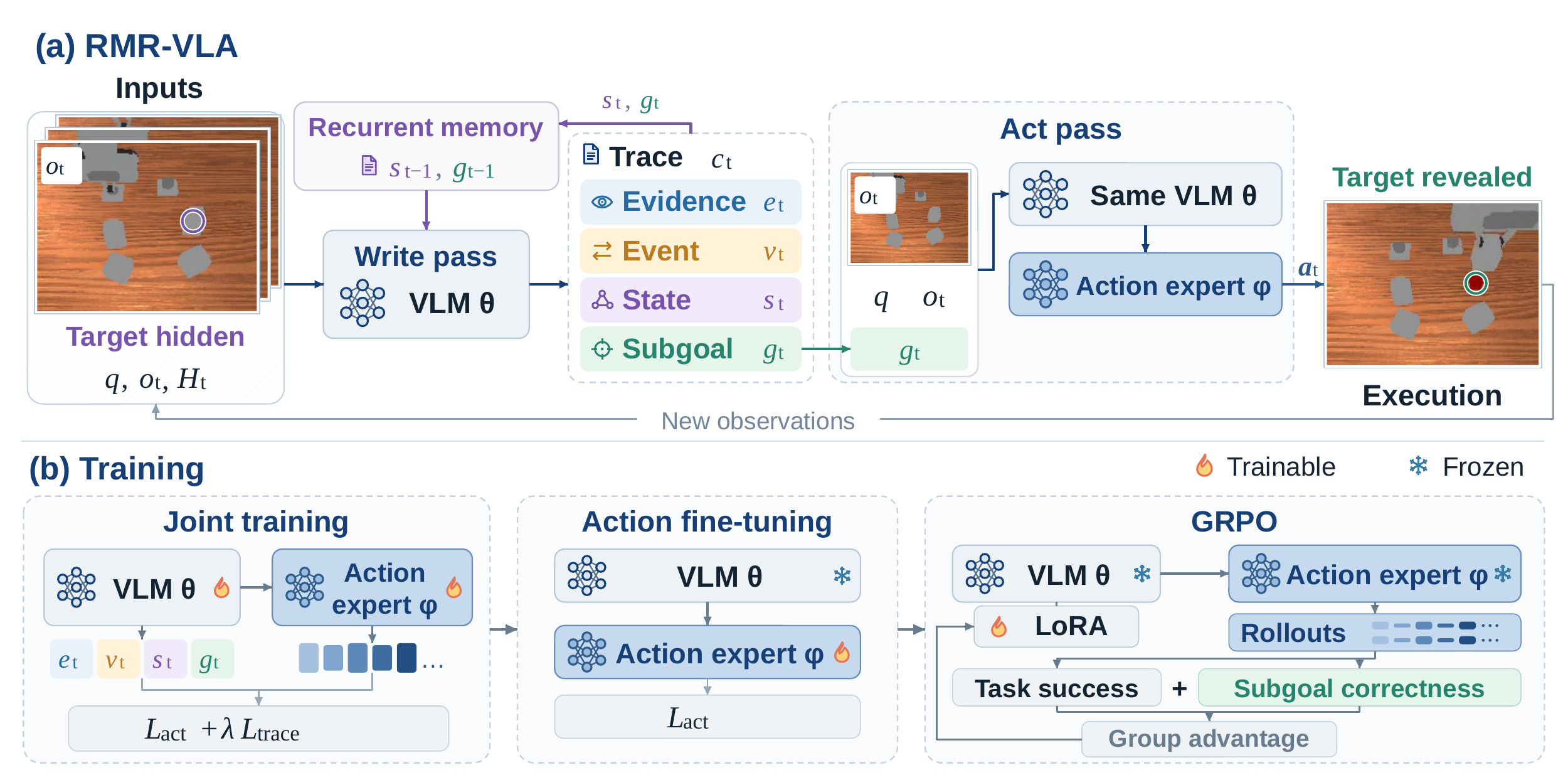}
    \caption{\textbf{Subgoals connect recurrent memory to action.}
    (a) In the write pass, the VLM writes the four-field trace; in the act pass, the same VLM encodes the current inputs with the grounded subgoal for the action expert.
    Only state and subgoal recur (purple); the subgoal conditions control (green).
    Camera views from a ButtonUnmaskSwap episode show the target hidden before execution and revealed after repeated action chunks.
    (b) Joint training learns reasoning and control; the VLM is then frozen while the action expert is fine-tuned.
    GRPO updates VLM LoRA with the base VLM weights and the action expert frozen, using task success and subgoal correctness.}
    \label{fig:loop}
\end{figure*}

\paragraph{Retaining and compressing visual history.}
Memory policies keep past observations through frame sampling~\cite{robomme}, token selection~\cite{divide}, feature memory banks~\cite{fang2025sam2act,memoryvla}, gated attention over cached tokens~\cite{gmp}, or, for long videos, adaptive memory management~\cite{jin2026videomem,zhu2025memorize}.
Within each step, token pruning guided by semantic attention compresses a VLA's visual input for efficient manipulation~\cite{ma2026safe}.
LTM-$\pi$ encodes visual transitions~\cite{ltm} and ActMem-VLA encodes executed actions~\cite{actmem}; other policies store episodic tokens, fast weights, object slots, or recurrent and retrieved visual streams~\cite{memoryvam,t2mem,slotvla,liberomem,dualstream}.
These representations supply temporal context for action prediction; \rmr{} turns it into an explicit task state and a grounded subgoal.

\paragraph{Embodied reasoning before action.}
ECoT predicts plans, subtasks, motion descriptions, and spatial features before action prediction~\cite{ecot}, and \pizero{} predicts a semantic subtask for its action expert~\cite{pi05}; other policies keep language records of plans, subtasks, or events across steps~\cite{notestoself,tacitvla,unimem}.
\rmr{} makes the subtask the result of recurrent memory reasoning: the current state resolves which target to act on, and the subgoal supplies it to control.

\paragraph{Memory and planning with external models.}
GAMMA coordinates external memory and reasoning agents~\cite{gamma}, MemER retrieves keyframes for an external VLM that instructs a low-level policy~\cite{sridhar2026memer}, and HeLM and 2AM steer low-level action models with external commands, task memory, or 2D hints~\cite{helm,twoam}; PAL and PonderPounce add planning or context models~\cite{pal,ponderpounce}.
LoopVLA stores completed subtasks as event memory, segmented by online Qwen3-VL-4B subgoals~\cite{zhu2026loopvla}.
\rmr{} learns memory reasoning and control in its shared VLM and action expert.

\paragraph{Reasoning in multimodal models.}
Explicit reasoning also strengthens multimodal models beyond manipulation: AdaThinkV adapts its thinking for token-efficient video reasoning~\cite{tian2026adathinkv}, Context-CoT synthesizes high-quality reasoning to enhance context learning~\cite{jin2026context}, OP-CAD distills on-policy from clean audio for robust audio-visual reasoning~\cite{shui2026op}, and Ponder \& Press advances visual GUI agents toward general computer control~\cite{wang2025ponder}.
Other multimodal models align disentangled audio semantics with visual features for audio-visual segmentation~\cite{tian2026delayed} or keep style aligned across generated images~\cite{zhang2025alignedgen}.
\rmr{} brings explicit reasoning into the memory of a VLA and refines the resulting trace with GRPO.

\section{Recurrent Memory Reasoning}
\label{sec:method}

\rmr{} maintains a current task state so that earlier visual cues continue to determine the robot's next target after they leave view (\cref{fig:loop}).
At each update, the VLM interprets new evidence, revises this state, and grounds the next subgoal; the action expert executes the subgoal, and the resulting observations feed the next update.

\subsection{Recurrent Memory Interface}
\label{sec:interface}

Let $t$ index memory updates, $q$ denote the task instruction, $o_t$ the current observation, comprising the camera views and the robot's proprioception, and $H_t$ the visual history, a bounded window of past frames, each pooled into a few visual tokens as in~\cite{chen2024videollm,robomme} (Appendix A).
\rmr{} keeps the \pizero{} architecture, a VLM with parameters $\theta$ and an action expert with parameters $\phi$; we write $F_\theta$ for the VLM and $F_{\theta,\phi}$ for the full model.
Each memory update consists of a write pass and an act pass.

\paragraph{Write pass.}
The VLM reads these inputs together with the recurrent pair $(s_{t-1},g_{t-1})$ and generates the \emph{trace}
\begin{equation}
    c_t = F_\theta(q,o_t,H_t,s_{t-1},g_{t-1})
        = [e_t,v_t,s_t,g_t],
    \label{eq:vlm}
\end{equation}
whose fields are the observation evidence $e_t$, the typed event $v_t$, the maintained state $s_t$, and the grounded subgoal $g_t$, produced as one autoregressive text output.

\paragraph{Act pass.}
The full model then predicts a continuous action chunk by flow matching~\cite{lipman2023flow,pi0}, conditioned on the subgoal just written:
\begin{equation}
    \mathbf{a}_t = F_{\theta,\phi}(q,o_t,g_t).
    \label{eq:action}
\end{equation}
The VLM encodes these inputs for the action expert.

\paragraph{The subgoal carries memory to control.}
The act pass conditions on the present inputs and $g_t$, while the visual history and the maintained state remain in the write pass.
Everything control needs from the past therefore reaches it through the subgoal.
The subgoal resolves which object to act on, such as the cup that now hides red, and guides the action expert's execution.
The recurrent pair $(s_t,g_t)$ carries the updated task information into the next write.

\subsection{Reasoning over the Current State}
\label{sec:reasoning}

\begin{table}[t]
    \centering
    \footnotesize
    \caption{\textbf{Trace fields in generation order.}
    Abridged model outputs for the \emph{Update} and \emph{Ground} columns of \cref{fig:teaser}.}
    \label{tab:trace}
    \setlength{\tabcolsep}{4pt}
    \renewcommand{\arraystretch}{1.12}
    \newcommand{\tracetext}[1]{{\scriptsize\ttfamily\frenchspacing #1}}
    \begin{tabularx}{\columnwidth}{@{}>{\raggedright\arraybackslash}p{0.265\columnwidth}YY@{}}
        \toprule
        Field & \emph{Update} & \emph{Ground} \\
        \midrule
        \textcolor{rmrevidence}{\textbf{Evidence}} $e_t$ \newline {\scriptsize\color{rmrgray}what the frames show}
            & \tracetext{moving red, blue}
            & \tracetext{done: pressed left button} \\
        \addlinespace[2pt]
        \textcolor{rmrevent}{\textbf{Event}} $v_t$ \newline {\scriptsize\color{rmrgray}what has changed}
            & \tracetext{swap (red, blue)}
            & \tracetext{none} \\
        \addlinespace[2pt]
        \textcolor{rmrstate}{\textbf{State}} $s_t$ \newline {\scriptsize\color{rmrgray}what holds now}
            & \tracetext{step 1; \dots; red <120,156>; blue <102,179>}
            & \tracetext{step 2; \dots; \textbf{red <94,168>}; blue <128,169>} \\
        \addlinespace[2pt]
        \textcolor{rmrgoal}{\textbf{Subgoal}} $g_t$ \newline {\scriptsize\color{rmrgray}what to do next}
            & \tracetext{press the left button at <66,99> until pressed}
            & \tracetext{pick up the container at \textbf{<94,168>} \dots} \\
        \bottomrule
    \end{tabularx}
\end{table}

Each write must leave behind a state that is still true after the scene has changed and hand the action expert a target it can act on now.
The trace does both with four fields written in the order in which they depend on one another (\cref{tab:trace}): the VLM commits to what it saw before it edits the state, and updates the state before it chooses a subgoal.

\paragraph{The state changes only when the scene gives a reason.}
The \emph{observation evidence} $e_t$ reports motion, contact, and completion cues.
The \emph{typed event} $v_t$ identifies a change such as covering, swapping, highlighting, or arrival at a target, with \emph{none} when no such event occurs.
The \emph{maintained state} $s_t$ lists task progress and the image location of each task-relevant object under that object's identity.
The VLM revises $s_{t-1}$ using both evidence and events: an event updates the affected identity--location associations or arrival count, completion evidence advances task progress, and unaffected entries carry over.
Completion cues can therefore advance progress while $v_t$ is \emph{none}.
A cube that disappears under a cup stays in the state as \emph{red}, at a location that follows its cup, and a cue shown once, such as the order of demonstrated placements, stays in the state after it has left the visual history.

\paragraph{The subgoal turns the state into a target.}
The \emph{grounded subgoal} $g_t$ resolves the instruction against the updated state, naming the operation, the image location of its target, and the condition that completes it.
The location is read from the state, so the action expert receives a spatially explicit target after the cue that singled it out has left view.
The completion condition serves the next write: with $g_{t-1}$ in its input, the VLM knows which operation is under way and what evidence finishes it.

\paragraph{Seen early, used late.}
In \cref{fig:teaser}, the red and blue cups trade places while the robot is pressing a button: the event moves red and blue in the state, while the subgoal is still to press the left button.
Four updates later, the button is pressed, and the new subgoal lifts the cup at the location the state now records for red (\cref{tab:trace}, bold).
The swap entered memory when it was seen and reached control when it was needed; the recurrent state carried it across the updates in between.

\subsection{Joint Reasoning and Control}
\label{sec:joint}

\paragraph{Trace labels.}
Trace labels are generated automatically: at each update of a training demonstration, annotation rules convert the simulator's object states and task progress into the four fields of \cref{sec:reasoning}, using only information available up to that update (Appendix A).

\paragraph{Joint objective.}
Both passes share the VLM, so we train them jointly with $\mathcal{L}=\mathcal{L}_{\mathrm{act}}+\lambda\mathcal{L}_{\mathrm{trace}}$.
The act pass learns flow matching on annotated action chunks:
\begin{equation}
    \mathcal{L}_{\mathrm{act}} = \mathbb{E}_{\tau,\boldsymbol{\epsilon}}\big\|v_{\theta,\phi}(\mathbf{a}^{\tau}_t,\tau\mid q,o_t,g^{\star}_t)-(\boldsymbol{\epsilon}-\mathbf{a}^{\star}_t)\big\|^2,
    \label{eq:act}
\end{equation}
where $\star$ marks annotated quantities, $v_{\theta,\phi}$ is the predicted velocity, and $\mathbf{a}^{\tau}_t=\tau\boldsymbol{\epsilon}+(1-\tau)\mathbf{a}^{\star}_t$ interpolates between the annotated chunk and noise $\boldsymbol{\epsilon}\sim\mathcal{N}(\mathbf{0},\mathbf{I})$.
The write pass learns the annotated trace by weighted token cross-entropy,
\begin{equation}
    \mathcal{L}_{\mathrm{trace}} = -\sum_{i} w_{t,i}\log p_\theta\big(c^{\star}_{t,i}\mid c^{\star}_{t,<i},\mathbf{x}_t\big),
    \label{eq:trace}
\end{equation}
where $\mathbf{x}_t=(q,o_t,H_t,s^{\star}_{t-1},g^{\star}_{t-1})$ collects the write-pass inputs and $w_{t,i}$ weights the $i$-th token.
The act pass trains on annotated subgoals and runs on generated ones; since the trace is discrete text, the two losses interact only through $\theta$.
Trace training uses selective token weighting and recoverable-history perturbations, inspired by prior work~\cite{kim2020somdst,wen2020copycat} (Appendix A).

\paragraph{Action-expert fine-tuning.}
After joint training, we freeze $\theta$ and fine-tune $\phi$ with $\mathcal{L}_{\mathrm{act}}$.
This preserves the learned trace generation while adapting continuous control to the fixed VLM representation and subgoal interface (\cref{sec:training}).

\subsection{Learning from Rollouts}
\label{sec:grpo}

\begin{table*}[t]
    \centering
    \caption{\textbf{RoboMME success by task (\%, $\uparrow$).} We evaluate 50 episodes per task; published scores come from the cited reports, with $^{*}$marking evaluations by RoboMME~\cite{robomme}. External reasoning models are high-level models outside the VLA backbone. Bold/underline mark the best/second-best within each group; dashes indicate unreported scores. Task abbreviations: Appendix C.}
    \label{tab:main}
    \scriptsize
    \setlength{\tabcolsep}{1.0pt}
    \begin{tabular*}{\textwidth}{@{\extracolsep{\fill}}l@{\hspace{8pt}}*{17}{r}@{}}
        \toprule
        Method & \multicolumn{4}{c}{Counting} & \multicolumn{4}{c}{Permanence} & \multicolumn{4}{c}{Reference} & \multicolumn{4}{c}{Imitation} & Avg. $\uparrow$ \\
        \cmidrule(lr){2-5}\cmidrule(lr){6-9}\cmidrule(lr){10-13}\cmidrule(lr){14-17}
         & BF & PX & SX & SC & VU & BU & VUS & BUS & PH & VR & VPB & VPO & MC & IP & PL & RS & \\
        \midrule
        \multicolumn{18}{@{}l}{\textit{With an external reasoning model}} \\
        MemER$^{*}$ (Qwen3-VL-4B)~\cite{sridhar2026memer} & 56.67 & 79.33 & 59.33 & 0.00 & 81.33 & 72.00 & 38.00 & 21.33 & 70.67 & 25.33 & 30.00 & 26.00 & \best{82.67} & \best{6.67} & 16.67 & 12.00 & 42.38 \\
        LoopVLA (Qwen3-VL-4B)~\cite{zhu2026loopvla} & 40.67 & \second{94.00} & \best{97.33} & \second{49.33} & 43.33 & 26.67 & 26.67 & 11.33 & 24.00 & 30.00 & 55.33 & 39.33 & \best{82.67} & \best{6.67} & \second{61.33} & \best{66.00} & 47.17 \\
        PonderPounce (Qwen3.5-9B)~\cite{ponderpounce} & 50.67 & 93.33 & 72.00 & \best{82.67} & \second{94.67} & \best{97.33} & 25.33 & \second{34.00} & \best{85.33} & 40.67 & \second{83.33} & \second{79.33} & \second{81.33} & \second{6.00} & 12.67 & 34.67 & \second{60.83} \\
        PAL (Qwen3-VL-4B)~\cite{pal} & \second{65.20} & 87.20 & \second{79.60} & 5.20 & \best{95.20} & \second{90.80} & 35.20 & 20.40 & 68.40 & 32.00 & 80.00 & \best{83.20} & 76.40 & 4.80 & 50.80 & 22.40 & 56.05 \\
        GAMMA (Qwen3.5-9B)~\cite{gamma} & \best{71.10} & \best{98.90} & 68.90 & 35.50 & 86.70 & 84.40 & \best{75.60} & \best{35.60} & \second{71.10} & \second{65.60} & 75.50 & 66.70 & 62.20 & 3.30 & \best{98.90} & 56.70 & \best{66.0} \\
        GAMMA (Qwen3.5-0.8B)~\cite{gamma} & 54.40 & 76.70 & 52.20 & 40.00 & 65.60 & 76.70 & \second{43.30} & 25.60 & 51.10 & \best{68.90} & \best{87.80} & 77.80 & 41.10 & 0.00 & \best{98.90} & \second{58.90} & 57.4 \\
        \midrule
        \multicolumn{18}{@{}l}{\textit{Without an external reasoning model}} \\
        \pizero{}~\cite{robomme} & 30.00 & 42.89 & 35.56 & 6.67 & 20.44 & 22.22 & 18.67 & 6.67 & 11.33 & 0.44 & 31.11 & 25.78 & 26.00 & 1.56 & 2.89 & 4.67 & 17.93 \\
        MemoryVLA$^{*}$~\cite{memoryvla} & 10.00 & 17.33 & 1.33 & 0.00 & 16.00 & 8.00 & 4.67 & 4.67 & 7.33 & 0.00 & 13.33 & 4.00 & 14.00 & 0.00 & 12.00 & 1.33 & 7.12 \\
        SAM2Act+$^{*}$~\cite{fang2025sam2act} & 40.00 & 76.00 & 25.33 & 0.00 & 27.33 & 32.00 & 18.00 & 26.67 & 17.33 & 5.33 & 24.67 & 20.00 & 29.33 & 0.00 & 0.00 & 0.00 & 21.37 \\
        FrameSamp+Modul~\cite{robomme} & 39.56 & 87.33 & 92.00 & 42.00 & 32.67 & 25.11 & 24.44 & 18.22 & 22.89 & 30.44 & 60.00 & 32.00 & 77.78 & 7.56 & 53.56 & 66.67 & 44.51 \\
        D\&R (256 tokens)~\cite{divide} & --- & --- & --- & --- & --- & --- & --- & --- & --- & --- & --- & --- & --- & --- & --- & --- & 50.11 \\
        Dual-Stream Memory~\cite{dualstream} & 42.00 & 93.33 & \second{92.67} & 55.33 & 35.00 & 28.00 & 23.00 & 25.00 & 32.67 & 35.00 & \second{74.00} & 39.56 & \second{83.33} & 8.00 & 52.00 & \second{68.67} & 49.22 \\
        T$^2$Mem~\cite{t2mem} & 60.67 & 91.33 & 90.67 & \second{70.00} & \second{88.00} & 59.33 & 29.33 & 20.00 & 50.67 & 38.67 & 35.33 & 30.00 & 80.67 & \best{36.00} & 50.00 & \best{78.67} & 56.83 \\
        LTM-$\pi$ (DreamDojo)~\cite{ltm} & 66.00 & \second{99.33} & \best{96.00} & \best{100.00} & 72.67 & \second{86.67} & \best{34.00} & 36.00 & 38.67 & 42.67 & 38.00 & 30.67 & 77.33 & \second{20.00} & \best{60.00} & 38.67 & 58.54 \\
        TacitVLA~\cite{tacitvla} & --- & --- & --- & --- & --- & --- & --- & --- & --- & --- & --- & --- & --- & --- & --- & --- & 48.1 \\
        \rmr{} (supervised) & \second{66.67} & \best{100.00} & 82.00 & 12.67 & 78.67 & 81.33 & \second{31.33} & \second{38.00} & \second{63.33} & \second{54.00} & \best{86.67} & \second{87.33} & 82.67 & 8.67 & 34.67 & 47.33 & \second{59.71} \\
        \rmr{} + GRPO & \best{73.33} & \second{99.33} & 92.00 & 6.00 & \best{94.67} & \best{93.33} & 30.67 & \best{50.67} & \best{69.33} & \best{72.67} & \best{86.67} & \best{90.67} & \best{90.67} & 5.33 & \second{56.67} & 51.33 & \best{66.46} \\
        \bottomrule
    \end{tabular*}
\end{table*}

Training on rollouts exposes the model to observations arising from its own decisions.
Flow matching gives no tractable action likelihood, so ReinFlow and $\pi$RL first make action generation stochastic by injecting noise into its denoising steps~\cite{reinflow,pirl}.
In \rmr{}, the decisions that carry memory to control are trace tokens with exact likelihoods.
Starting from the supervised model, we therefore apply GRPO~\cite{shao2024deepseekmath,guo2025deepseekr1} directly to the generated trace, updating VLM LoRA~\cite{hu2022lora} while freezing the action expert and its flow-matching sampler.
The episode reward combines task completion with subgoal correctness:
\begin{equation}
    R = r_{\mathrm{term}} + \beta m.
    \label{eq:reward}
\end{equation}
The terminal term $r_{\mathrm{term}}$ is $+1$ for success and $-1$ for failure or timeout; the subgoal correctness $m$ is the fraction of the subgoal decisions generated in the episode, one per update, whose text and coordinates agree with the correct labels.
A group of rollouts with the same terminal outcome gives no learning signal under $r_{\mathrm{term}}$ alone~\cite{tian2026adathinkv}; $m$ distinguishes them.
For each task layout we sample $K$ episodes; episode $k$ receives reward $R_k$ and generates $n_k$ trace tokens $c_{k,j}$ over its updates.
We maximize the GRPO objective
\begin{equation}
    \mathcal{J}_{\mathrm{GRPO}} = \frac{1}{K}\sum_{k=1}^{K}\frac{1}{n_k}\sum_{j=1}^{n_k}\min\Big(\rho_{k,j}A_k,
    \mathrm{clip}\big(\rho_{k,j},1-\delta,1+\delta\big)A_k\Big),
    \label{eq:grpo}
\end{equation}
with the token probability ratio and the leave-one-out advantage~\cite{kool2019buy,ahmadian2024back}
\begin{equation}
    \rho_{k,j} = \frac{p_\theta(c_{k,j}\mid\cdot)}{p_{\theta_{\mathrm{old}}}(c_{k,j}\mid\cdot)},\quad A_k = R_k-\frac{1}{K-1}\sum_{k'\neq k}R_{k'},
    \label{eq:advantage}
\end{equation}
where both probabilities condition on the inputs and preceding tokens of the same update.
A KL constraint regularizes the update (Appendix B).

\section{Experiments}
\label{sec:experiments}

We compare \rmr{} with existing memory methods on long-horizon, history-dependent robotic manipulation (\cref{sec:results,sec:liberomem}), trace its gains to the trace fields and their recurrence (\cref{sec:structure}), and examine how a single VLA learns to write the trace and act on it (\cref{sec:training,sec:feedback_results}).

\subsection{Evaluation Setup}
\label{sec:setup}

\paragraph{Benchmark and training data.}
RoboMME contains 16 tasks in four categories of four~\cite{robomme}: Counting tracks repeated operations, Permanence tracks objects through occlusion and swaps, Reference requires acting on an earlier observation, and Imitation requires reproducing a demonstration.
Supervised training uses the official demonstrations, which we also annotate with trace labels.
At evaluation, the model maintains its own state and generates its own subgoals.
We also train and evaluate on LIBERO-Mem~\cite{liberomem}, whose ten tasks require remembering an object's last motion, how many times an operation has been repeated, how identical bowls have been rearranged, and which of two identical baskets holds an object (Appendix C).

\paragraph{Metrics.}
Following the official protocols, we evaluate 50 episodes per task on RoboMME and 40 on LIBERO-Mem, averaging success over tasks and three evaluation seeds.

\subsection{Acting on Past Evidence}
\label{sec:results}

\begin{table*}[t]
    \centering
    \caption{\textbf{LIBERO-Mem success by task (\%, $\uparrow$).} Each method is listed at its best reported average with the matching task scores. We evaluate 40 episodes per task. $^{\dagger}$Relaxed success: completing the full ordered goal before further repetition~\cite{twoam}. Bold/underline mark the best/second-best within each group; dashes indicate unreported scores. Task definitions: Appendix C.}
    \label{tab:liberomem}
    \footnotesize
    \setlength{\tabcolsep}{2.6pt}
    \begin{tabular*}{\textwidth}{@{\extracolsep{\fill}}>{\raggedright\arraybackslash}p{.26\textwidth}*{11}{r}@{}}
        \toprule
        Method & \multicolumn{2}{c}{Motion} & \multicolumn{4}{c}{Sequence} & \multicolumn{2}{c}{Relation} & \multicolumn{2}{c}{Occlusion} & Avg. $\uparrow$ \\
        \cmidrule(lr){2-3}\cmidrule(lr){4-7}\cmidrule(lr){8-9}\cmidrule(lr){10-11}
         & T1 & T2 & T3 & T4 & T5 & T6 & T7 & T8 & T9 & T10 & \\
        \midrule
        \multicolumn{12}{@{}l}{\textit{With an external reasoning model}} \\
        2AM (Qwen3.8-27B)~\cite{twoam}$^{\dagger}$ & \best{95.83} & \best{100.00} & \best{89.17} & \best{75.00} & \best{85.83} & \best{90.83} & \second{10.00} & \best{10.00} & \second{47.50} & \second{25.83} & \best{63.00} \\
        HeLM (Qwen2.5-VL-7B)~\cite{helm} & \second{75.00} & \second{60.00} & \second{75.00} & \second{55.00} & \second{65.00} & \second{25.00} & \best{45.00} & \best{10.00} & \best{85.00} & \best{90.00} & \second{58.5} \\
        \midrule
        \multicolumn{12}{@{}l}{\textit{Without an external reasoning model}} \\
        $\pi_0$ (reproduced by 2AM)~\cite{pi0}$^{\dagger}$ & 71.67 & 91.67 & 52.50 & 40.00 & 42.50 & 54.17 & 6.67 & 9.17 & 0.00 & 5.83 & 37.42 \\
        \pizero{}~\cite{pi05,actmem} & 76.70 & 90.00 & 51.70 & 40.00 & 81.70 & 35.00 & 25.00 & 51.70 & \best{100.00} & \best{100.00} & 65.2 \\
        MemoryVLA~\cite{memoryvla,actmem} & \second{98.30} & 88.30 & 15.00 & 20.00 & 3.30 & 0.00 & 33.30 & 53.30 & 91.70 & 91.70 & 49.5 \\
        GMP~\cite{gmp,memoryvam} & 85.00 & 45.00 & 20.00 & 5.00 & 40.00 & 0.00 & 0.00 & 0.00 & 0.00 & 0.00 & 19.5 \\
        MemoryVAM (UNet)~\cite{memoryvam} & 85.00 & 65.00 & 90.00 & 10.00 & 35.00 & 0.00 & 25.00 & 5.00 & 40.00 & 70.00 & 42.5 \\
        MemoryVAM (DiT)~\cite{memoryvam} & 75.00 & 65.00 & 65.00 & 15.00 & 55.00 & 35.00 & 10.00 & 10.00 & 25.00 & 50.00 & 40.5 \\
        LTM-$\pi$ (CD-LAM)~\cite{ltm} & 95.00 & 75.00 & 90.00 & 70.00 & 70.00 & 45.00 & \best{80.00} & 20.00 & \best{100.00} & \best{100.00} & 74.5 \\
        LTM-$\pi$ (DreamDojo)~\cite{ltm} & 95.00 & 75.00 & \best{100.00} & \best{90.00} & 90.00 & 65.00 & 65.00 & 25.00 & \best{100.00} & \best{100.00} & 80.5 \\
        ActMem-VLA~\cite{actmem} & 93.30 & \second{92.80} & 81.10 & 85.60 & 95.00 & 75.60 & 33.30 & \second{55.00} & 97.20 & \second{98.90} & 80.8 \\
        SmoLSTM~\cite{habekost2026smolstm} & --- & --- & --- & --- & --- & --- & --- & --- & --- & --- & 77.5 \\
        STVLA~\cite{han2026stvla} & 95.00 & 0.00 & 90.00 & 0.00 & 70.00 & 40.00 & 5.00 & 5.00 & 25.00 & 80.00 & 41.0 \\
        \rmr{} (supervised) & 95.83 & \best{95.00} & \best{100.00} & 85.83 & \second{95.83} & \second{95.00} & \second{72.50} & 51.67 & \best{100.00} & \best{100.00} & \second{89.17} \\
        \rmr{} + GRPO & \best{99.17} & 92.50 & \second{99.17} & \second{87.50} & \best{98.33} & \best{97.50} & 64.17 & \best{63.33} & \second{99.17} & \best{100.00} & \best{90.08} \\
        \bottomrule
    \end{tabular*}
\end{table*}

\input{content/figures/rmr_libero_returns}
\input{content/figures/rmr_trace_ablation}

\paragraph{Earlier cues decide later targets.}
In the four Reference tasks, an earlier observation determines the target of a later action.
\rmr{} + GRPO reaches 79.83\% on them, the highest Reference average among the compared methods, including systems with an external reasoning model (\cref{tab:main}).
The margin is widest on VideoPlaceOrder, whose targets follow the order of demonstrated placements: \rmr{} + GRPO reaches 90.67\%, against 39.56\% for the strongest published policy without an external reasoning model.
The retained order guides target selection after the demonstration has left view (\cref{fig:case_study}a).

\paragraph{Overall success in a single VLA.}
Across all 16 tasks, \rmr{} + GRPO reaches 66.46\%, the best overall result among the compared methods: 7.92 points above the best method without an external reasoning model, LTM-$\pi$~\cite{ltm} (DreamDojo~\cite{gao2026dreamdojo}), and above the best system with one, GAMMA with a 9B agent (66.0\%).

\paragraph{Where the trace helps less.}
\rmr{} trails the strongest memory policies on StopCube (6.00\%, against 100\% for LTM-$\pi$), which requires pressing a button at the moment a moving cube reaches its target on a specified pass.
The cube passes the target between updates, so the press must land within one action chunk, a timing precision that the trace does not control.

\subsection{Object Memory on LIBERO-Mem}
\label{sec:liberomem}

On LIBERO-Mem (\cref{tab:liberomem}), supervised \rmr{} reaches 89.17\% and GRPO raises it to 90.08\%; the best published results, ActMem-VLA (80.8\%) and LTM-$\pi$ (DreamDojo, 80.5\%), trail the supervised model by over eight points.

\begin{table*}[t]
    \centering
    \footnotesize
    \caption{\textbf{Trace ablation on RoboMME (\%, $\uparrow$).} All variants are trained without GRPO. \emph{Without recurrence} writes all four fields without recurrent input; recurrent traces cumulatively add fields to $g_t$. Correctness is subgoal correctness $m$ (\cref{eq:reward}). Bold marks column maxima.}
    \label{tab:structure}
    \renewcommand{\arraystretch}{1.08}
    \setlength{\tabcolsep}{3.0pt}
    \begin{tabular*}{\textwidth}{@{\extracolsep{\fill}}lcccccrrrrr@{}}
        \toprule
         & \multicolumn{5}{c}{Subgoal diagnostics} & \multicolumn{5}{c}{Task success} \\
        \cmidrule(lr){2-6}\cmidrule(l){7-11}
        Trace & Count/stop & Container & Cue target & Demo step & Correctness & Count. & Perm. & Ref. & Imit. & Overall \\
        \textit{Task group} & \textit{Counting} & \textit{Permanence} & \textit{Reference} & \textit{Imitation} & \textit{All} & & & & & \\
        \midrule
        Without recurrence & 66.0 & 30.0 & 22.5 & 43.2 & 42.6 & 30.17 & 35.17 & 32.83 & 20.83 & 29.75 \\
        \midrule
        Subgoal $g_t$ only & 73.2 & 43.5 & 38.5 & 54.8 & 54.0 & 29.50 & 42.50 & 32.17 & 24.17 & 32.08 \\
        + state $s_t$ & 81.7 & 67.0 & 78.7 & 69.2 & 72.1 & 54.83 & 44.50 & 58.67 & 35.67 & 48.42 \\
        + evidence $e_t$ & 86.5 & 68.0 & 78.5 & 72.1 & 74.3 & 62.67 & 52.17 & 67.17 & 41.00 & 55.75 \\
        + event $v_t$ (\rmr{}) & \best{92.2} & \best{71.5} & \best{82.8} & \best{75.3} & \best{77.7} & \best{65.33} & \best{57.33} & \best{72.83} & \best{43.33} & \best{59.71} \\
        \bottomrule
    \end{tabular*}
\end{table*}

\begin{figure*}[t]
    \centering
    \includegraphics[width=\textwidth]{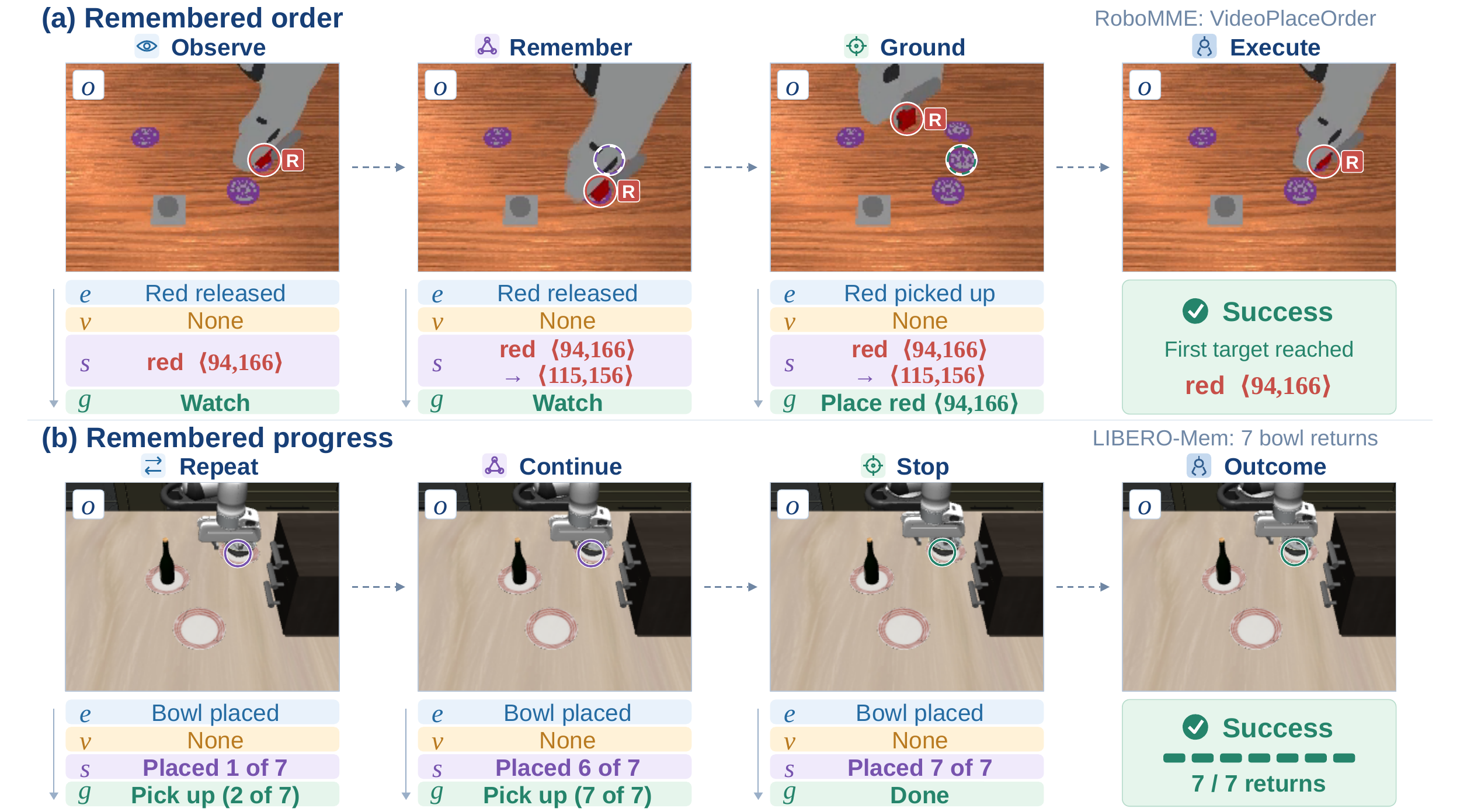}
    \caption{\textbf{Memory guides targets and progress.}
    \rmr{} rollouts on RoboMME (a) and LIBERO-Mem (b), with views ($o$) and abbreviated traces ($e,v,s,g$).
    In (a), red labels track the cube; coordinates $\langle y,x\rangle$ denote remembered placements.
    Dashed arrows skip intervening updates; outcome panels show completion.}
    \label{fig:case_study}
\end{figure*}

\paragraph{Success holds as bowl returns accumulate.}
Across the bowl-return tasks requiring one, three, five, and seven returns, \rmr{} + GRPO stays at or above 97.50\% (\cref{fig:returns}), while no policy without an external reasoning model reports over 75.6\% at seven returns.
Each completed return advances a count in the state, so the seventh return poses the same update as the first; in \cref{fig:case_study}b, near-identical views carry different counts, and the count decides whether the next subgoal picks up again or ends the task.

\paragraph{Tracking identical objects.}
The two relation tasks rearrange visually identical bowls, the demand illustrated by the cup swap in \cref{fig:teaser}.
\rmr{} + GRPO obtains the highest reported score for rotating three bowls (63.33\%), and supervised \rmr{} reaches 72.50\% for swapping two, second only to LTM-$\pi$ (CD-LAM~\cite{wei2026cdlam}, 80.0\%; \cref{tab:liberomem}).
These are also the two lowest LIBERO-Mem tasks for both models, just as the two swap tasks are their lowest in RoboMME Permanence: following identical objects through a rearrangement remains the hardest update for the state.

\subsection{Trace and Recurrence}
\label{sec:structure}

We vary which fields the trace writes and whether state and subgoal recur, measuring subgoal correctness alongside task success (\cref{tab:structure,fig:structure}).

\paragraph{Subgoal diagnostics.}
Each diagnostic compares generated subgoals with the correct labels: \emph{count/stop} and \emph{demo step} check each decision's text, \emph{container} the identity and location at each episode's first cup choice, and \emph{cue target} the text and location of decisions that act on earlier cues.
Subgoal correctness $m$, the per-episode fraction of correct subgoal decisions (\cref{eq:reward}), is averaged per task and then equally over the 16 tasks (Appendix C).

\paragraph{Recurrence carries progress and earlier cues forward.}
With the same four fields supervised, recurrence of $(s_{t-1},g_{t-1})$ raises subgoal correctness $m$ from 42.6\% to 77.7\% and task success from 29.75\% to 59.71\%, improving all 16 tasks.
Reference and Counting gain the most, 40.00 and 35.17 points: their targets depend on a cue shown earlier or on how many operations have been completed.

\paragraph{The maintained state is what recurrence must carry.}
A recurrent subgoal alone reaches 32.08\%, against 29.75\% without recurrence.
Adding the state $s_t$ raises success to 48.42\% and brings the largest diagnostic gains of any field, 40.2 points in cue target and 23.5 in container (\cref{fig:structure}a); within Reference, VideoPlaceButton and VideoPlaceOrder gain the most (\cref{fig:structure}b).
A bare recurrent subgoal names the current target, but Counting and Reference need the progress and earlier cues that the state keeps across updates.

\paragraph{Evidence and events sharpen each update.}
Evidence raises count/stop correctness by 4.8 points and success to 55.75\%; the event raises all four diagnostics and success to 59.71\%.
Permanence success rises mainly in these two steps, from 44.50\% to 57.33\%: ButtonUnmask gains 16.67 points with evidence, which reports completions such as a pressed button, and VideoUnmaskSwap gains 10.00 with the event, which names the swap (\cref{fig:structure}b).

\subsection{Training Schedules}
\label{sec:training}

\begin{table}[t]
    \centering
    \footnotesize
    \caption{\textbf{Training schedules on RoboMME (\%, $\uparrow$).} All use equal training steps and no GRPO. Sequential training optimizes the trace loss, then the action loss; with two models, separate models were trained to generate traces and actions, respectively. Joint training optimizes both losses throughout or ends with action-expert fine-tuning (supervised \rmr{}). All seven schedules: Appendix C. Bold marks column maxima.}
    \label{tab:training_main}
    \setlength{\tabcolsep}{2.6pt}
    \renewcommand{\arraystretch}{1.08}
    \begin{tabular}{@{}lrrrrr@{}}
        \toprule
        Schedule & Count. & Perm. & Ref. & Imit. & Avg. $\uparrow$ \\
        \midrule
        \multicolumn{6}{@{}l}{\textit{Sequential training}} \\
        Single model & 5.50 & 9.17 & 9.33 & 9.17 & 8.29 \\
        Single model, frozen VLM & 14.17 & 37.67 & 42.67 & 15.50 & 27.50 \\
        Two models & 37.17 & 48.33 & 69.67 & 33.00 & 47.04 \\
        \midrule
        \multicolumn{6}{@{}l}{\textit{Joint training}} \\
        Throughout & 58.50 & 50.33 & 71.83 & 38.83 & 54.88 \\
        Then action-expert fine-tuning & \best{65.33} & \best{57.33} & \best{72.83} & \best{43.33} & \best{59.71} \\
        \bottomrule
    \end{tabular}
\end{table}

\paragraph{Training the roles in sequence loses one of them.}
Under the same number of training steps and without GRPO (\cref{tab:training_main}), sequential training optimizes the trace loss, then the action loss.
If the action stage also updates the VLM, the model loses the trace it acts on: task success falls to 8.29\%, and rises to 47.04\% when separate models generate traces and actions, at the cost of running two models.
With a frozen VLM, the trace survives, but the action expert uses a representation never shaped by control (27.50\%).

\paragraph{Joint training keeps both; fine-tuning the action expert sharpens control.}
Optimizing both losses throughout reaches 54.88\% with a single model, above the two-model pipeline.
Ending with action-expert fine-tuning instead, with the VLM frozen, raises success to 59.71\% with gains in all four categories: once joint training has settled what the VLM writes and represents, the action expert refines control on a fixed interface.
This schedule is the best of the seven in every category (Appendix C).

\subsection{Rollout Feedback}
\label{sec:feedback_results}

\begin{table}[t]
    \centering
    \footnotesize
    \caption{\textbf{RL ablation on RoboMME (\%, $\uparrow$).} $\pi$RL-style trains only the action expert; GRPO only VLM LoRA. Episode reward adds subgoal correctness $m$ (\cref{eq:reward}) to the terminal reward. Bold marks column maxima.}
    \label{tab:rl}
    \setlength{\tabcolsep}{2.6pt}
    \renewcommand{\arraystretch}{1.08}
    \begin{tabular}{@{}lrrrrr@{}}
        \toprule
        Configuration & Count. & Perm. & Ref. & Imit. & Avg. $\uparrow$ \\
        \midrule
        \rmr{} (supervised) & 65.33 & 57.33 & 72.83 & 43.33 & 59.71 \\
        \midrule
        $\pi$RL-style, terminal reward & 57.33 & 43.67 & 72.83 & 43.67 & 54.38 \\
        GRPO, terminal reward & 65.33 & 57.83 & 76.17 & 46.17 & 61.38 \\
        GRPO, episode reward & \best{67.67} & \best{67.33} & \best{79.83} & \best{51.00} & \best{66.46} \\
        \bottomrule
    \end{tabular}
\end{table}

$\pi$RL-style training~\cite{pirl}, which updates only the action expert with terminal reward, lowers success from 59.71\% to 54.38\%.
GRPO on the generated traces, with the action expert frozen, instead raises it to 61.38\% with terminal reward alone and to 66.46\% with subgoal correctness $m$ added (\cref{tab:rl}), the latter improving all four category averages and 11 of the 16 tasks.
The lift from $m$ is largest on Permanence (9.50 points), where each correct subgoal depends on tracking a hidden object through the episode.

\FloatBarrier
\section{Conclusion}
\label{sec:conclusion}

RMR lets a vision-language-action model remember by reasoning.
Before each action chunk, the VLA writes a text trace: it revises an explicit task state from new evidence and grounds the subgoal its action expert executes, so a cue enters memory when it is seen and reaches control when it is needed.
Carrying the state and subgoal forward doubles RoboMME success over an otherwise identical model without recurrence, and among the trace fields the state brings the largest gain in subgoal correctness.
Joint training keeps reasoning and control in one VLA, and GRPO refines the text trace from rollout rewards with the action expert frozen.
By writing, revising, and acting on its own memory, \rmr{} achieves the best RoboMME and LIBERO-Mem results among the compared methods.

{\small
\bibliographystyle{plainnat}
\bibliography{content/references}

\begin{thebibliography}{50}
\providecommand{\natexlab}[1]{#1}
\providecommand{\url}[1]{\texttt{#1}}
\expandafter\ifx\csname urlstyle\endcsname\relax
  \providecommand{\doi}[1]{doi: #1}\else
  \providecommand{\doi}{doi: \begingroup \urlstyle{rm}\Url}\fi

\bibitem[Ahmadian et~al.(2024)Ahmadian, Cremer, Gall{\'e}, Fadaee, Kreutzer,
  Pietquin, {\"U}st{\"u}n, and Hooker]{ahmadian2024back}
Arash Ahmadian, Chris Cremer, Matthias Gall{\'e}, Marzieh Fadaee, Julia
  Kreutzer, Olivier Pietquin, Ahmet {\"U}st{\"u}n, and Sara Hooker.
\newblock {Back to Basics: Revisiting REINFORCE-Style Optimization for Learning
  from Human Feedback in LLMs}.
\newblock In Lun-Wei Ku, Andre Martins, and Vivek Srikumar, editors,
  \emph{Proceedings of the 62nd Annual Meeting of the Association for
  Computational Linguistics (Volume 1: Long Papers)}, pages 12248--12267,
  Bangkok, Thailand, August 2024. Association for Computational Linguistics.
\newblock \doi{10.18653/v1/2024.acl-long.662}.
\newblock URL \url{https://aclanthology.org/2024.acl-long.662/}.

\bibitem[Black et~al.(2024)Black, Brown, Driess, Esmail, Equi, Finn, Fusai,
  Groom, Hausman, Ichter, Jakubczak, Jones, Ke, Levine, Li-Bell, Mothukuri,
  Nair, Pertsch, Shi, Tanner, Vuong, Walling, Wang, and Zhilinsky]{pi0}
Kevin Black, Noah Brown, Danny Driess, Adnan Esmail, Michael Equi, Chelsea
  Finn, Niccolo Fusai, Lachy Groom, Karol Hausman, Brian Ichter, Szymon
  Jakubczak, Tim Jones, Liyiming Ke, Sergey Levine, Adrian Li-Bell, Mohith
  Mothukuri, Suraj Nair, Karl Pertsch, Lucy~Xiaoyang Shi, James Tanner, Quan
  Vuong, Anna Walling, Haohuan Wang, and Ury Zhilinsky.
\newblock {$\pi_0$: A Vision-Language-Action Flow Model for General Robot
  Control}.
\newblock \emph{arXiv preprint arXiv:2410.24164}, 2024.
\newblock URL \url{https://arxiv.org/abs/2410.24164}.

\bibitem[Chen et~al.(2024)Chen, Lv, Wu, Lin, Song, Gao, Liu, Gao, Mao, and
  Shou]{chen2024videollm}
Joya Chen, Zhaoyang Lv, Shiwei Wu, Kevin~Qinghong Lin, Chenan Song, Difei Gao,
  Jia-Wei Liu, Ziteng Gao, Dongxing Mao, and Mike~Zheng Shou.
\newblock {VideoLLM-online: Online Video Large Language Model for Streaming
  Video}.
\newblock In \emph{Proceedings of the IEEE/CVF Conference on Computer Vision
  and Pattern Recognition}, pages 18407--18418, 2024.
\newblock URL \url{https://arxiv.org/abs/2406.11816}.

\bibitem[Chen et~al.(2025)Chen, Liu, Zhang, Guo, Xu, Lin, Zang, Li, Wei, Zhou,
  Zhang, Yu, Fan, Huang, Wang, and Yu]{pirl}
Kang Chen, Zhihao Liu, Tonghe Zhang, Zhen Guo, Si~Xu, Hao Lin, Hongzhi Zang,
  Xiang Li, Bingwen Wei, Jiakai Zhou, Quanlu Zhang, Zhaofei Yu, Guoliang Fan,
  Tiejun Huang, Yu~Wang, and Chao Yu.
\newblock {$\pi_{\texttt{RL}}$: Online RL Fine-tuning for Flow-based
  Vision-Language-Action Models}.
\newblock \emph{arXiv preprint arXiv:2510.25889}, 2025.
\newblock URL \url{https://arxiv.org/abs/2510.25889}.

\bibitem[Choi et~al.(2026)Choi, Jung, Kim, Lee, and Yu]{ponderpounce}
Suhwan Choi, Jaeyoon Jung, Sungkyung Kim, Yunsung Lee, and Youngjae Yu.
\newblock {PonderPounce: A Pretrained MLLM as an Episode Context Engine for
  Robot Control}.
\newblock \emph{arXiv preprint arXiv:2608.24115}, 2026.
\newblock URL \url{https://arxiv.org/abs/2608.24115}.

\bibitem[Chung et~al.(2025)Chung, Hanyu, Nguyen, Le, Bumgarner, Nguyen, Vo,
  Yamazaki, Rainwater, Kieu, Nguyen, and Le]{liberomem}
Nhat Chung, Taisei Hanyu, Toan Nguyen, Huy Le, Frederick Bumgarner, Duy Minh~Ho
  Nguyen, Khoa Vo, Kashu Yamazaki, Chase Rainwater, Tung Kieu, Anh Nguyen, and
  Ngan Le.
\newblock {Rethinking Progression of Memory State in Robotic Manipulation: An
  Object-Centric Perspective}.
\newblock \emph{arXiv preprint arXiv:2511.11478}, 2025.
\newblock URL \url{https://arxiv.org/abs/2511.11478}.

\bibitem[Dai et~al.(2026)Dai, Fu, Lee, Liu, Zhang, Yang, Finn, Fazeli, and
  Chai]{robomme}
Yinpei Dai, Hongze Fu, Jayjun Lee, Yuejiang Liu, Haoran Zhang, Jianing Yang,
  Chelsea Finn, Nima Fazeli, and Joyce Chai.
\newblock {RoboMME: Benchmarking and Understanding Memory for Robotic
  Generalist Policies}.
\newblock \emph{arXiv preprint arXiv:2603.04639}, 2026.
\newblock URL \url{https://arxiv.org/abs/2603.04639}.

\bibitem[Du et~al.(2026)Du, Guo, Zhang, Wang, Dong, Gao, Han, Lu, Yu, Shen,
  Chang, and Li]{sera}
Tingwen Du, Haoyu Guo, Hanxue Zhang, Xuanxu Wang, Junting Dong, Chao Gao,
  Mingfei Han, Tao Lu, Mulin Yu, Chunhua Shen, Xiaojun Chang, and Zhihui Li.
\newblock {From VLM Planners to Robotic Agents: Self-Evolving Harnesses for
  History-Dependent Manipulation}.
\newblock \emph{arXiv preprint arXiv:2610.29560}, 2026.
\newblock URL \url{https://arxiv.org/abs/2610.29560}.

\bibitem[Fang et~al.(2025)Fang, Grotz, Pumacay, Wang, Fox, Krishna, and
  Duan]{fang2025sam2act}
Haoquan Fang, Markus Grotz, Wilbert Pumacay, Yi~Ru Wang, Dieter Fox, Ranjay
  Krishna, and Jiafei Duan.
\newblock {SAM2Act: Integrating Visual Foundation Model with A Memory
  Architecture for Robotic Manipulation}.
\newblock In \emph{Proceedings of the 42nd International Conference on Machine
  Learning}, volume 267 of \emph{Proceedings of Machine Learning Research},
  pages 15925--15942, 2025.
\newblock URL \url{https://proceedings.mlr.press/v267/fang25c.html}.

\bibitem[Gao et~al.(2026{\natexlab{a}})Gao, Liang, Zheng, Malik, Ye, Yu, Tseng,
  Dong, Mo, Lin, Ma, Nah, Magne, Xiang, Xie, Zheng, Niu, Tan, Zentner, Kurian,
  Indupuru, Jannaty, Gu, Zhang, Malik, Abbeel, Liu, Zhu, Jang, and
  Fan]{gao2026dreamdojo}
Shenyuan Gao, William Liang, Kaiyuan Zheng, Ayaan Malik, Seonghyeon Ye, Sihyun
  Yu, Wei-Cheng Tseng, Yuzhu Dong, Kaichun Mo, Chen-Hsuan Lin, Qianli Ma,
  Seungjun Nah, Loic Magne, Jiannan Xiang, Yuqi Xie, Ruijie Zheng, Dantong Niu,
  You~Liang Tan, K.~R. Zentner, George Kurian, Suneel Indupuru, Pooya Jannaty,
  Jinwei Gu, Jun Zhang, Jitendra Malik, Pieter Abbeel, Ming-Yu Liu, Yuke Zhu,
  Joel Jang, and Linxi Fan.
\newblock {DreamDojo: A Generalist Robot World Model from Large-Scale Human
  Videos}.
\newblock \emph{arXiv preprint arXiv:2602.06949}, 2026{\natexlab{a}}.
\newblock URL \url{https://arxiv.org/abs/2602.06949}.

\bibitem[Gao et~al.(2026{\natexlab{b}})Gao, Liu, Li, and Song]{gmp}
Yihuai Gao, Jeff~Jinyun Liu, Shuang Li, and Shuran Song.
\newblock {Gated Memory Policy: In-Context Memorization and Adaptation}.
\newblock \emph{arXiv preprint arXiv:2604.18933}, 2026{\natexlab{b}}.
\newblock URL \url{https://arxiv.org/abs/2604.18933}.

\bibitem[Guo et~al.(2025)Guo, Yang, Zhang, et~al.]{guo2025deepseekr1}
Daya Guo, Dejian Yang, Haowei Zhang, et~al.
\newblock {DeepSeek-R1 incentivizes reasoning in LLMs through reinforcement
  learning}.
\newblock \emph{Nature}, 645\penalty0 (8081):\penalty0 633--638, 2025.
\newblock \doi{10.1038/s41586-025-09422-z}.
\newblock URL \url{https://doi.org/10.1038/s41586-025-09422-z}.

\bibitem[Habekost et~al.(2026)Habekost, Kashani, G{\"a}de, Kerzel, Allgeuer,
  Weber, Wermter, and Lee]{habekost2026smolstm}
Jan-Gerrit Habekost, Parsa~Mastouri Kashani, Connor G{\"a}de, Matthias Kerzel,
  Philipp Allgeuer, Cornelius Weber, Stefan Wermter, and Jae~Hee Lee.
\newblock {SmoLSTM: A Compact Vision-Language-Action Model with Recurrent
  Memory that Persists}.
\newblock \emph{arXiv preprint arXiv:2609.22854}, 2026.
\newblock URL \url{https://arxiv.org/abs/2609.22854}.

\bibitem[Han et~al.(2026)Han, Wang, Wang, Han, Xiong, Wan, Mao, Peng, Liu,
  Shan, and Yao]{han2026stvla}
Zhuo Han, Yeqiang Wang, Xinran Wang, Hongwei Han, Yi~Xiong, Zunian Wan, Jianbo
  Mao, Huan Peng, Yongcheng Liu, Wei Shan, and Maoqing Yao.
\newblock {STVLA: Spatial-aware and Temporal-memory for Vision-Language-Action
  Models}.
\newblock \emph{arXiv preprint arXiv:2611.16589}, 2026.
\newblock URL \url{https://arxiv.org/abs/2611.16589}.

\bibitem[Hanyu et~al.(2025)Hanyu, Chung, Le, Nguyen, Ikebe, Gunderman, Minh,
  Vo, Kieu, Yamazaki, Rainwater, Nguyen, and Le]{slotvla}
Taisei Hanyu, Nhat Chung, Huy Le, Toan Nguyen, Yuki Ikebe, Anthony Gunderman,
  Duy Nguyen~Ho Minh, Khoa Vo, Tung Kieu, Kashu Yamazaki, Chase Rainwater, Anh
  Nguyen, and Ngan Le.
\newblock {SlotVLA: Towards Modeling of Object-Relation Representations in
  Robotic Manipulation}.
\newblock \emph{arXiv preprint arXiv:2511.06754}, 2025.
\newblock URL \url{https://arxiv.org/abs/2511.06754}.

\bibitem[Haresh et~al.(2026)Haresh, Dijkman, Bhattacharyya, and
  Memisevic]{notestoself}
Sanjay Haresh, Daniel Dijkman, Apratim Bhattacharyya, and Roland Memisevic.
\newblock {Notes-to-Self: Scratchpad Augmented VLAs for Memory Dependent
  Manipulation Tasks}.
\newblock \emph{arXiv preprint arXiv:2602.21013}, 2026.
\newblock URL \url{https://arxiv.org/abs/2602.21013}.

\bibitem[Hu et~al.(2022)Hu, Shen, Wallis, Allen-Zhu, Li, Wang, Wang, and
  Chen]{hu2022lora}
Edward~J. Hu, Yelong Shen, Phillip Wallis, Zeyuan Allen-Zhu, Yuanzhi Li, Shean
  Wang, Lu~Wang, and Weizhu Chen.
\newblock {LoRA: Low-Rank Adaptation of Large Language Models}.
\newblock In \emph{International Conference on Learning Representations}, 2022.
\newblock URL \url{https://openreview.net/forum?id=nZeVKeeFYf9}.

\bibitem[Hu et~al.(2026)Hu, Chen, Cao, and Detry]{twoam}
Yutong Hu, Fengjiao Chen, Xuezhi Cao, and Renaud Detry.
\newblock {2AM: Grounding Agent-Side Memory as Guidance for Steerable Action
  Models in Long-Horizon Manipulation}.
\newblock \emph{arXiv preprint arXiv:2609.11308}, 2026.
\newblock URL \url{https://arxiv.org/abs/2609.11308}.

\bibitem[Jiang et~al.(2026)Jiang, Yu, Chen, Feng, Yang, Gite, and
  Jiang]{memoryvam}
Yuxin Jiang, Chang Yu, Yunuo Chen, Xiang Feng, Yin Yang, Nishank Gite, and
  Chenfanfu Jiang.
\newblock {MemoryVAM: Integrating Memory into Video Action Model for Robot
  Manipulation}.
\newblock \emph{arXiv preprint arXiv:2606.20679}, 2026.
\newblock URL \url{https://arxiv.org/abs/2606.20679}.

\bibitem[Jin et~al.(2026{\natexlab{a}})Jin, Tian, Ding, Zhang, Zhang, Xie, Liu,
  and Cheng]{jin2026videomem}
Hongbo Jin, Jingqi Tian, Jiayu Ding, Qiaoman Zhang, Wenhao Zhang, Siyi Xie,
  Yang Liu, and Sijie Cheng.
\newblock {VideoMem: Enhancing Ultra-Long Video Understanding via Adaptive
  Memory Management}.
\newblock In \emph{Findings of the Association for Computational Linguistics:
  EMNLP 2026}, pages 2066--2081, 2026{\natexlab{a}}.

\bibitem[Jin et~al.(2026{\natexlab{b}})Jin, Zhu, Tian, Jiang, Du, Tang, Xie,
  Zhang, and Ding]{jin2026context}
Hongbo Jin, Mingnan Zhu, Jingqi Tian, Xu~Jiang, Zhongjing Du, Haoran Tang, Siyi
  Xie, Qiaoman Zhang, and Jiayu Ding.
\newblock {Context-CoT: Enhancing Context Learning via High-Quality Reasoning
  Synthesis}.
\newblock \emph{arXiv preprint arXiv:2605.25354}, 2026{\natexlab{b}}.
\newblock URL \url{https://arxiv.org/abs/2605.25354}.

\bibitem[Kim et~al.(2026)Kim, Kim, Choi, and Lee]{helm}
Gyeonghun Kim, Donghoon Kim, Minsoo Choi, and Suyun Lee.
\newblock {HeLM: Hierarchical Explicit Language Memory for Non-Markovian
  Manipulation}.
\newblock \emph{arXiv preprint arXiv:2611.02660}, 2026.
\newblock URL \url{https://arxiv.org/abs/2611.02660}.

\bibitem[Kim et~al.(2020)Kim, Yang, Kim, and Lee]{kim2020somdst}
Sungdong Kim, Sohee Yang, Gyuwan Kim, and Sang-Woo Lee.
\newblock {Efficient Dialogue State Tracking by Selectively Overwriting
  Memory}.
\newblock In Dan Jurafsky, Joyce Chai, Natalie Schluter, and Joel Tetreault,
  editors, \emph{Proceedings of the 58th Annual Meeting of the Association for
  Computational Linguistics}, pages 567--582, Online, July 2020. Association
  for Computational Linguistics.
\newblock \doi{10.18653/v1/2020.acl-main.53}.
\newblock URL \url{https://aclanthology.org/2020.acl-main.53/}.

\bibitem[Kool et~al.(2019)Kool, van Hoof, and Welling]{kool2019buy}
Wouter Kool, Herke van Hoof, and Max Welling.
\newblock {Buy 4 REINFORCE Samples, Get a Baseline for Free!}
\newblock In \emph{ICLR Workshop on Deep Reinforcement Learning Meets
  Structured Prediction}, 2019.
\newblock URL \url{https://openreview.net/forum?id=r1lgTGL5DE}.

\bibitem[Lipman et~al.(2023)Lipman, Chen, Ben-Hamu, Nickel, and
  Le]{lipman2023flow}
Yaron Lipman, Ricky T.~Q. Chen, Heli Ben-Hamu, Maximilian Nickel, and Matt Le.
\newblock {Flow Matching for Generative Modeling}.
\newblock In \emph{International Conference on Learning Representations}, 2023.
\newblock URL \url{https://openreview.net/forum?id=PqvMRDCJT9t}.

\bibitem[Liu et~al.(2026{\natexlab{a}})Liu, Wang, Li, He, Gao, Wang, Sun, Liu,
  and Liu]{pal}
Qinghang Liu, Conglin Wang, Zongsheng Li, Boyong He, Han Gao, Pengcheng Wang,
  Tao Sun, Lige Liu, and Chang Liu.
\newblock {Plan First, Act Later: Streaming Reasoning for Executable Memory in
  Robot Control}.
\newblock \emph{arXiv preprint arXiv:2610.29329}, 2026{\natexlab{a}}.
\newblock URL \url{https://arxiv.org/abs/2610.29329}.

\bibitem[Liu et~al.(2026{\natexlab{b}})Liu, Huang, Hong, Du, Cao, Fei-Fei, and
  Wu]{t2mem}
Yize Liu, Huang Huang, Yining Hong, Zijian Du, Zhi Cao, Li~Fei-Fei, and Jiajun
  Wu.
\newblock {T$^2$Mem: Learning Test-Time Memory for Robotics}.
\newblock \emph{arXiv preprint arXiv:2609.36720}, 2026{\natexlab{b}}.
\newblock URL \url{https://arxiv.org/abs/2609.36720}.

\bibitem[Liu et~al.(2026{\natexlab{c}})Liu, Lin, Tang, Zheng, Li, Zhang, Li,
  Jing, Liu, Gao, Bi, Sun, Chen, and Li]{gamma}
Yudong Liu, Yueqian Lin, Zijia Tang, Yuxi Zheng, Yuan Li, Shuai Zhang, Yi~Li,
  Taotao Jing, Shuangjun Liu, Dashan Gao, Ning Bi, Jingwei Sun, Yiran Chen, and
  Hai~Helen Li.
\newblock {Gamma: Grounded Multi-Agent Memory with Mechanical Audit for
  Memory-Intensive Manipulation}.
\newblock \emph{arXiv preprint arXiv:2610.12793}, 2026{\natexlab{c}}.
\newblock URL \url{https://arxiv.org/abs/2610.12793}.

\bibitem[Lu et~al.(2026)Lu, You, Li, Cao, Tian, Xiao, Chen, Luo, Ivanovic,
  Pavone, and Wang]{tacitvla}
Jack Lu, Yurong You, Boyi Li, Yulong Cao, Thomas Tian, Chaowei Xiao, Xiangyu
  Chen, Wenjie Luo, Boris Ivanovic, Marco Pavone, and Yan Wang.
\newblock {TacitVLA: Reasoning Vision-Language-Action Models Without Reasoning
  Latency}.
\newblock \emph{arXiv preprint arXiv:2610.17540}, 2026.
\newblock URL \url{https://arxiv.org/abs/2610.17540}.

\bibitem[Ma et~al.(2026)Ma, Zhang, Wang, Wang, Wu, Wang, Tian, Zhu, and
  Tang]{ma2026safe}
Shilin Ma, Chubin Zhang, Changyuan Wang, Yuji Wang, Yue Wu, Zixuan Wang, Jingqi
  Tian, Zheng Zhu, and Yansong Tang.
\newblock {SAFE-Pruner: Semantic Attention--Guided Future-Aware Token Pruning
  for Efficient Vision-Language-Action Manipulation}.
\newblock In \emph{European Conference on Computer Vision}, pages 163--180.
  Springer, 2026.

\bibitem[Osterberg et~al.(2026)Osterberg, Wang, and Schwager]{unimem}
Lars Osterberg, Maggie Wang, and Mac Schwager.
\newblock {UniMem: Unifying Visual and Textual Memory in VLAs for Non-Markovian
  Tasks}.
\newblock \emph{arXiv preprint arXiv:2608.22869}, 2026.
\newblock URL \url{https://arxiv.org/abs/2608.22869}.

\bibitem[{Physical Intelligence} et~al.(2025){Physical Intelligence}, Black,
  Brown, Darpinian, Dhabalia, Driess, Esmail, Equi, Finn, Fusai, Galliker,
  Ghosh, Groom, Hausman, Ichter, Jakubczak, Jones, Ke, LeBlanc, Levine,
  Li-Bell, Mothukuri, Nair, Pertsch, Ren, Shi, Smith, Springenberg, Stachowicz,
  Tanner, Vuong, Walke, Walling, Wang, Yu, and Zhilinsky]{pi05}
{Physical Intelligence}, Kevin Black, Noah Brown, James Darpinian, Karan
  Dhabalia, Danny Driess, Adnan Esmail, Michael Equi, Chelsea Finn, Niccolo
  Fusai, Manuel~Y. Galliker, Dibya Ghosh, Lachy Groom, Karol Hausman, Brian
  Ichter, Szymon Jakubczak, Tim Jones, Liyiming Ke, Devin LeBlanc, Sergey
  Levine, Adrian Li-Bell, Mohith Mothukuri, Suraj Nair, Karl Pertsch, Allen~Z.
  Ren, Lucy~Xiaoyang Shi, Laura Smith, Jost~Tobias Springenberg, Kyle
  Stachowicz, James Tanner, Quan Vuong, Homer Walke, Anna Walling, Haohuan
  Wang, Lili Yu, and Ury Zhilinsky.
\newblock {$\pi_{0.5}$: a Vision-Language-Action Model with Open-World
  Generalization}.
\newblock \emph{arXiv preprint arXiv:2504.16054}, 2025.
\newblock URL \url{https://arxiv.org/abs/2504.16054}.

\bibitem[Shao et~al.(2024)Shao, Wang, Zhu, Xu, Song, Bi, Zhang, Zhang, Li, Wu,
  and Guo]{shao2024deepseekmath}
Zhihong Shao, Peiyi Wang, Qihao Zhu, Runxin Xu, Junxiao Song, Xiao Bi, Haowei
  Zhang, Mingchuan Zhang, Y.~K. Li, Y.~Wu, and Daya Guo.
\newblock {DeepSeekMath: Pushing the Limits of Mathematical Reasoning in Open
  Language Models}.
\newblock \emph{arXiv preprint arXiv:2402.03300}, 2024.
\newblock URL \url{https://arxiv.org/abs/2402.03300}.

\bibitem[Shi et~al.(2025)Shi, Xie, Liu, Sun, Liu, Wang, Zhou, Fan, Zhang, and
  Huang]{memoryvla}
Hao Shi, Bin Xie, Yingfei Liu, Lin Sun, Fengrong Liu, Tiancai Wang, Erjin Zhou,
  Haoqiang Fan, Xiangyu Zhang, and Gao Huang.
\newblock {MemoryVLA: Perceptual-Cognitive Memory in Vision-Language-Action
  Models for Robotic Manipulation}.
\newblock \emph{arXiv preprint arXiv:2508.19236}, 2025.
\newblock URL \url{https://arxiv.org/abs/2508.19236}.

\bibitem[Shui et~al.(2026)Shui, Chen, Liu, Tian, Li, Yi, Hong, and
  Tang]{shui2026op}
Xingming Shui, Dapeng Chen, Bowei Liu, Jingqi Tian, Minfu Li, Kun Yi, Jiapeng
  Hong, and Yansong Tang.
\newblock {OP-CAD: On-Policy Clean-Audio Distillation for Robust Audio-Visual
  Reasoning}.
\newblock \emph{arXiv preprint arXiv:2609.39150}, 2026.
\newblock URL \url{https://arxiv.org/abs/2609.39150}.

\bibitem[Sridhar et~al.(2026)Sridhar, Pan, Sharma, and Finn]{sridhar2026memer}
Ajay Sridhar, Jennifer Pan, Satvik Sharma, and Chelsea Finn.
\newblock {MemER: Scaling Up Memory for Robot Control via Experience
  Retrieval}.
\newblock In \emph{International Conference on Learning Representations}, 2026.
\newblock URL \url{https://arxiv.org/abs/2510.20328}.

\bibitem[Sun et~al.(2026)Sun, Gao, Chen, Wang, Liu, and Wang]{ltm}
Qixin Sun, WenLong Gao, Xiansheng Chen, Zhongyuan Wang, Si~Liu, and Pengwei
  Wang.
\newblock {LTM: Latent Transition Based In-Context Memory for
  Vision-Language-Action Models}.
\newblock \emph{arXiv preprint arXiv:2610.21362}, 2026.
\newblock URL \url{https://arxiv.org/abs/2610.21362}.

\bibitem[Tian et~al.(2026{\natexlab{a}})Tian, Du, Zhang, Wang, Lee, Bai, Zhu,
  Niu, and Tang]{tian2026delayed}
Jingqi Tian, Yiheng Du, Haoji Zhang, Yuji Wang, Isaac~Ning Lee, Xulong Bai,
  Tianrui Zhu, Jingxuan Niu, and Yansong Tang.
\newblock {Delayed Bidirectional Alignment via Disentangled Audio Semantics for
  Audio-Visual Segmentation}.
\newblock In \emph{European Conference on Computer Vision}, pages 619--638.
  Springer, 2026{\natexlab{a}}.

\bibitem[Tian et~al.(2026{\natexlab{b}})Tian, Zhang, Chen, Jin, Xu, Zhu, Shui,
  Ma, Yang, and Tang]{tian2026adathinkv}
Jingqi Tian, Haoji Zhang, Lin Chen, Hongbo Jin, Haonan Xu, Tianrui Zhu,
  Xingming Shui, Shilin Ma, Wenjing Yang, and Yansong Tang.
\newblock {AdaThinkV: Adaptive Thinking for Token-Efficient Video Reasoning}.
\newblock \emph{arXiv preprint arXiv:2608.01980}, 2026{\natexlab{b}}.
\newblock URL \url{https://arxiv.org/abs/2608.01980}.

\bibitem[Wang et~al.(2026)Wang, Liu, Wang, Chen, Zhang, and Zhang]{dualstream}
Yang Wang, Ping Liu, Ruiyu Wang, Xuhua Chen, Lingfeng Zhang, and Tao Zhang.
\newblock {Dual-Stream Memory for Vision-Language-Action Policies}.
\newblock \emph{arXiv preprint arXiv:2610.13262}, 2026.
\newblock URL \url{https://arxiv.org/abs/2610.13262}.

\bibitem[Wang et~al.(2025)Wang, Zhang, Tian, and Tang]{wang2025ponder}
Yiqin Wang, Haoji Zhang, Jingqi Tian, and Yansong Tang.
\newblock {Ponder \& Press: Advancing Visual GUI Agent towards General Computer
  Control}.
\newblock In \emph{Findings of the Association for Computational Linguistics:
  ACL 2025}, pages 1461--1473, 2025.

\bibitem[Wei et~al.(2026)Wei, Zhou, Mao, Xu, Liang, Zhang, Xi, Han, Yan, Wang,
  Feng, and Huang]{wei2026cdlam}
Yufan Wei, Kun Zhou, Lingjun Mao, Ziming Xu, Shuang Liang, Zijun Zhang, Ziqiao
  Xi, Ruobing Han, Yuchen Yan, Xinyue Wang, Fan Feng, and Biwei Huang.
\newblock {Causally Debiased Latent Action Model for Embodied
  Action-Conditioned World Models}.
\newblock \emph{arXiv preprint arXiv:2607.09185}, 2026.
\newblock URL \url{https://arxiv.org/abs/2607.09185}.

\bibitem[Wen et~al.(2020)Wen, Lin, Darrell, Jayaraman, and Gao]{wen2020copycat}
Chuan Wen, Jierui Lin, Trevor Darrell, Dinesh Jayaraman, and Yang Gao.
\newblock {Fighting Copycat Agents in Behavioral Cloning from Observation
  Histories}.
\newblock In H.~Larochelle, M.~Ranzato, R.~Hadsell, M.~F. Balcan, and H.~Lin,
  editors, \emph{Advances in Neural Information Processing Systems}, volume~33,
  pages 2564--2575. Curran Associates, Inc., 2020.
\newblock URL
  \url{https://proceedings.neurips.cc/paper/2020/hash/1b113258af3968aaf3969ca67e744ff8-Abstract.html}.

\bibitem[Yu et~al.(2026)Yu, Yu, Wang, Du, Albrecht, and Soh]{divide}
Xuehui Yu, Eason Yu, Meiyi Wang, Haozhe Du, Stefano~V. Albrecht, and Harold
  Soh.
\newblock {Divide-and-Remember: Recursive Action-Relevant Memory for
  Long-Horizon VLA Policies}.
\newblock \emph{arXiv preprint arXiv:2610.00982}, 2026.
\newblock URL \url{https://arxiv.org/abs/2610.00982}.

\bibitem[Zawalski et~al.(2024)Zawalski, Chen, Pertsch, Mees, Finn, and
  Levine]{ecot}
Micha{\l} Zawalski, William Chen, Karl Pertsch, Oier Mees, Chelsea Finn, and
  Sergey Levine.
\newblock {Robotic Control via Embodied Chain-of-Thought Reasoning}.
\newblock \emph{arXiv preprint arXiv:2407.08693}, 2024.
\newblock URL \url{https://arxiv.org/abs/2407.08693}.

\bibitem[Zhang et~al.(2025{\natexlab{a}})Zhang, Du, Wang, Li, Gu, and
  Zhang]{zhang2025alignedgen}
Jiexuan Zhang, Yiheng Du, Qian Wang, Weiqi Li, Yu~Gu, and Jian Zhang.
\newblock {AlignedGen: Aligning Style Across Generated Images}.
\newblock In D.~Belgrave, C.~Zhang, H.~Lin, R.~Pascanu, P.~Koniusz,
  M.~Ghassemi, and N.~Chen, editors, \emph{Advances in Neural Information
  Processing Systems}, volume~38, pages 98168--98192. Curran Associates, Inc.,
  2025{\natexlab{a}}.
\newblock \doi{10.52202/085713-3280}.
\newblock URL
  \url{https://proceedings.neurips.cc/paper_files/paper/2025/file/8d31937cf16463ca456d48b9768335b9-Paper-Conference.pdf}.

\bibitem[Zhang et~al.(2025{\natexlab{b}})Zhang, Yu, Su, and Wang]{reinflow}
Tonghe Zhang, Chao Yu, Sichang Su, and Yu~Wang.
\newblock {ReinFlow}: Fine-tuning flow matching policy with online
  reinforcement learning.
\newblock In \emph{Advances in Neural Information Processing Systems},
  2025{\natexlab{b}}.
\newblock URL \url{https://arxiv.org/abs/2505.22094}.

\bibitem[Zhao et~al.(2026)Zhao, Huang, Wang, Yang, and Jiang]{actmem}
Yaxin Zhao, Dianye Huang, Chenwei Wang, Chenguang Yang, and Zhongliang Jiang.
\newblock {ActMem-VLA: Action-History Memory with Dual-Expert Denoising for
  Long-Horizon Vision-Language-Action Policies}.
\newblock \emph{arXiv preprint arXiv:2609.37307}, 2026.
\newblock URL \url{https://arxiv.org/abs/2609.37307}.

\bibitem[Zhu et~al.(2026)Zhu, Liu, Wang, Wang, Yue, Liu, Wu, Xue, and
  Zeng]{zhu2026loopvla}
Shoujing Zhu, Zhenyang Liu, Fungmiu Wang, Jiafeng Wang, Bo~Yue, Guiliang Liu,
  Simo Wu, Xiangyang Xue, and Taiping Zeng.
\newblock {LoopVLA: Cross-Subtask Event Memory for Looped Task Execution}.
\newblock \emph{arXiv preprint arXiv:2606.17463}, 2026.
\newblock URL \url{https://arxiv.org/abs/2606.17463}.

\bibitem[Zhu et~al.(2025)Zhu, Zhang, Sun, Tian, and Tang]{zhu2025memorize}
Tianrui Zhu, Shiyi Zhang, Zhirui Sun, Jingqi Tian, and Yansong Tang.
\newblock {Memorize-and-Generate: Towards Long-Term Consistency in Real-Time
  Video Generation}.
\newblock \emph{arXiv preprint arXiv:2512.18741}, 2025.
\newblock URL \url{https://arxiv.org/abs/2512.18741}.

\end{thebibliography}
}

\clearpage
\appendix
\crefalias{section}{appendix}
\crefalias{subsection}{subappendix}
\suppressfloats[t]
\newcommand{\tracecell}[1]{{\scriptsize\ttfamily\frenchspacing #1}}
\newlength{\exunit}
\newcommand{\exframe}[7]{%
    \def\exleft{#2}\def\extop{#5}%
    \setlength{\exunit}{\dimexpr\linewidth/#6\relax}%
    \begin{tikzpicture}[baseline=(current bounding box.north),x=\exunit,y=-\exunit]
        \node[anchor=north west,inner sep=0pt] at (0,0)
            {#1};
        #7
    \end{tikzpicture}}
\newcommand{\exring}[3]{%
    \draw[white,line width=1.5pt] ({#3-\exleft},{#2-\extop}) circle[radius=3.1pt];
    \draw[#1,line width=0.75pt] ({#3-\exleft},{#2-\extop}) circle[radius=3.1pt];}

\section{Method and Training Details}
\label{app:method}

\subsection{Inputs and Update Schedule}
\label{app:inputs}

Every 16 frames, the write pass reads the instruction, the current camera views and proprioception, the visual history, and the preceding state and subgoal $(s_{t-1},g_{t-1})$.
The visual history covers the preceding 32 frames. Following FrameSamp in RoboMME~\cite{robomme} and VideoLLM-online~\cite{chen2024videollm}, we pool each frame into 16 visual tokens, for a total of 512.
The state and subgoal carry information beyond this window as text.
In the act pass, the shared VLM encodes the instruction, current observation, and the operation and target specified by the new subgoal. The robot executes 16 steps of the predicted action chunk before the next update.
The completion condition stays in the recurrent subgoal so that the next write pass can recognize when the operation is complete.

On RoboMME tasks with a demonstration video, the write pass runs at the same interval during playback and writes the subgoal \texttt{watch}.
At the first update after playback, the VLM uses the maintained state to identify the target for the first act pass (\cref{fig:example_repick}).
On LIBERO-Mem, each task begins with robot interaction, and an act pass follows every write pass.

\paragraph{Update interval and visual history.}
\Cref{tab:updatewindow} compares the update interval and the number of frames available to each write pass.
At intervals of 4, 8, and 16 frames, extending the history from one interval to two raises overall RoboMME success from 44.00\% to 45.79\%, from 43.29\% to 49.63\%, and from 57.96\% to 59.71\%, respectively.
With two intervals of history, the write pass sees frames on both sides of the preceding update.
For a fixed 32-frame history, shortening the interval from 32 to 16 frames lets the model revise its subgoal after 16 executed steps and raises success from 54.17\% to 59.71\%.

\begin{table*}[t]
    \centering
    \caption{\textbf{Update interval and visual history on RoboMME (success, \%).} The interval sets the number of frames between write passes and the number of action steps executed per act pass. The history length sets the number of preceding frames available to the write pass. Bold marks column maxima.}
    \label{tab:updatewindow}
    \scriptsize
    \setlength{\tabcolsep}{1.2pt}
    \renewcommand{\arraystretch}{1.12}
    \begin{tabular*}{\textwidth}{@{\extracolsep{\fill}}>{\raggedright\arraybackslash}p{.24\textwidth}*{17}{r}@{}}
        \toprule
        Interval / history (frames) & \multicolumn{4}{c}{Counting} & \multicolumn{4}{c}{Permanence} & \multicolumn{4}{c}{Reference} & \multicolumn{4}{c}{Imitation} & Avg. $\uparrow$ \\
        \cmidrule(lr){2-5}\cmidrule(lr){6-9}\cmidrule(lr){10-13}\cmidrule(lr){14-17}
         & BF & PX & SX & SC & VU & BU & VUS & BUS & PH & VR & VPB & VPO & MC & IP & PL & RS & \\
        \midrule
        4 / 4 & 32.00 & 61.33 & 41.33 & 6.67 & 70.00 & 82.00 & 23.33 & 23.33 & 57.33 & 27.33 & 79.33 & 78.67 & 63.33 & 2.00 & 26.00 & 30.00 & 44.00 \\
        4 / 8 & 52.00 & 71.33 & 51.33 & 8.67 & 51.33 & 70.67 & 21.33 & 38.00 & 69.33 & 32.00 & 68.67 & 64.00 & 62.00 & 2.00 & 30.00 & 40.00 & 45.79 \\
        8 / 8 & 38.67 & 68.67 & 73.33 & 6.00 & 48.00 & 80.00 & 14.00 & 24.00 & 67.33 & 27.33 & 62.67 & 56.67 & 63.33 & 2.67 & 28.00 & 32.00 & 43.29 \\
        8 / 16 & 50.00 & 73.33 & 79.33 & 18.67 & 53.33 & 77.33 & 25.33 & 30.67 & 67.33 & 44.00 & 71.33 & 72.67 & 68.00 & 2.00 & 30.00 & 30.67 & 49.63 \\
        16 / 16 & \best{67.33} & 95.33 & \best{85.33} & 15.33 & 59.33 & \best{86.00} & 20.67 & \best{46.67} & 65.33 & \best{56.00} & 85.33 & 84.00 & 82.00 & 5.33 & 34.00 & 39.33 & 57.96 \\
        16 / 32 (\rmr{}) & 66.67 & \best{100.00} & 82.00 & 12.67 & \best{78.67} & 81.33 & \best{31.33} & 38.00 & 63.33 & 54.00 & \best{86.67} & \best{87.33} & \best{82.67} & \best{8.67} & 34.67 & \best{47.33} & \best{59.71} \\
        32 / 32 & 60.67 & 91.33 & 43.33 & \best{22.67} & 75.33 & 76.00 & 24.67 & 43.33 & \best{76.00} & 53.33 & 83.33 & 84.67 & 72.00 & 7.33 & \best{39.33} & 13.33 & 54.17 \\
        \bottomrule
    \end{tabular*}
\end{table*}

\subsection{Trace Labels}
\label{app:labels}

At each update of a training demonstration, annotation rules use the simulator's object states and task progress up to that point to label the four trace fields in generation order.
\begin{itemize}[leftmargin=*,nosep]
    \item \textbf{Observation evidence $e_t$} reports contact and motion since the preceding update (\texttt{grasped}, \texttt{moving}, \texttt{released}), completion of the preceding subgoal (\texttt{done:}), or a demonstrated step (\texttt{shown:}). When the field reports none of these cues, its label is \texttt{still}.
    \item \textbf{Typed event $v_t$} names a change from a fixed vocabulary, such as covering, swapping, highlighting, or arrival at a target. An occurrence index (\texttt{\#n}) distinguishes repeated events. The label names an event once its completion is visible and reads \texttt{none} otherwise.
    \item \textbf{Maintained state $s_t$} carries forward object identities, locations, and task progress. An event updates the affected identity--location associations or arrival count. Completion evidence advances the step counter (\texttt{step k}) and the relevant progress counts. Other entries carry over. The rules update an object's location when it is held or moves more than about 2\,cm from its previous resting position. They mark an object in the gripper as \texttt{held} and record a resting position after about three consecutive frames in which the object is still.
    \item \textbf{Grounded subgoal $g_t$} specifies the next operation, its target and image location, and a completion condition (\texttt{until \dots}). Ordinals such as \emph{the second green cube} or \emph{for the first time} distinguish repeated operations and identify which instance a completion cue refers to. The subgoal matches the one the simulator assigns at that frame.
\end{itemize}

\Cref{tab:labelexamples} illustrates how evidence and events revise different parts of the state.
In BinFill, contact marks the green cube as \texttt{held} at step~0. At the next update, pickup completion advances the counter to step~1 and switches the subgoal to placing the cube in the bin. The event is \texttt{none} throughout.
In PickHighlight, a single highlighting event adds three cube identities and locations to the state. The entries persist at the next update while the subgoal remains to press the button.

\begin{table*}[t]
    \centering
    \footnotesize
    \caption{\textbf{Trace labels at three consecutive updates.}
    Each column shows the annotated trace at one update in a BinFill episode (top) and a PickHighlight episode (bottom). The state and subgoal from one column serve as inputs to the next.}
    \label{tab:labelexamples}
    \setlength{\tabcolsep}{3pt}
    \renewcommand{\arraystretch}{1.1}
    \begin{tabularx}{\textwidth}{@{}>{\raggedright\arraybackslash}p{0.075\textwidth}YYY@{}}
        \toprule
        & \multicolumn{1}{c}{$t$} & \multicolumn{1}{c}{$t+1$} & \multicolumn{1}{c}{$t+2$} \\
        \midrule
        \multicolumn{4}{@{}l}{\textit{BinFill}} \\
        \textcolor{rmrevidence}{\textbf{Evidence}}
            & \tracecell{still}
            & \tracecell{grasped green cube 4}
            & \tracecell{done: picked up green cube 4} \\
        \textcolor{rmrevent}{\textbf{Event}}
            & \tracecell{none}
            & \tracecell{none}
            & \tracecell{none} \\
        \textcolor{rmrstate}{\textbf{State}}
            & \tracecell{step 0; button <64,168>; bin <124,95>; green cube 4 <109,185>; green 0 of 3 in bin; blue 0 of 1 in bin}
            & \tracecell{step 0; button <64,168>; bin <124,95>; green cube 4 <109,185> held; green 0 of 3 in bin; blue 0 of 1 in bin}
            & \tracecell{step 1; button <64,168>; bin <124,95>; green 0 of 3 in bin; blue 0 of 1 in bin} \\
        \textcolor{rmrgoal}{\textbf{Subgoal}}
            & \tracecell{pick up the first green cube at <109,185> until picked up}
            & \tracecell{pick up the first green cube at <109,185> until picked up}
            & \tracecell{put it into the bin at <124,95> until put down} \\
        \midrule
        \multicolumn{4}{@{}l}{\textit{PickHighlight}} \\
        \textcolor{rmrevidence}{\textbf{Evidence}}
            & \tracecell{still}
            & \tracecell{still}
            & \tracecell{still} \\
        \textcolor{rmrevent}{\textbf{Event}}
            & \tracecell{none}
            & \tracecell{highlight\#1 blue cube 1, green cube 2, blue cube 4}
            & \tracecell{none} \\
        \textcolor{rmrstate}{\textbf{State}}
            & \tracecell{step 0; button <57,138>}
            & \tracecell{step 0; button <57,138>; blue cube 1 <56,100>; green cube 2 <80,126>; blue cube 4 <112,173>; highlighted: blue cube 1, green cube 2, blue cube 4}
            & \tracecell{step 0; button <57,138>; blue cube 1 <56,100>; green cube 2 <80,126>; blue cube 4 <112,173>; highlighted: blue cube 1, green cube 2, blue cube 4} \\
        \textcolor{rmrgoal}{\textbf{Subgoal}}
            & \tracecell{press the button at <57,138> until pressed}
            & \tracecell{press the button at <57,138> until pressed}
            & \tracecell{press the button at <57,138> until pressed} \\
        \bottomrule
    \end{tabularx}
\end{table*}

\subsection{Selective Token Weighting and Recoverable-History Perturbations}
\label{app:techniques}

Most entries in the state and subgoal carry over between consecutive traces (\cref{tab:labelexamples}).
During supervised training, the write pass receives the annotated pair $(s^{\star}_{t-1},g^{\star}_{t-1})$ as part of $\mathbf{x}_t$ (Eq.~\eqref{eq:trace}).
Selective token weighting emphasizes the entries that change, and recoverable-history perturbations train the VLM to use current evidence to correct a preceding entry.

\paragraph{Selective token weighting.}
In $\mathcal{L}_{\mathrm{trace}}$, tokens copied from the preceding state or subgoal and formatting tokens receive weight $w_{t,i}=1/3$.
Tokens for new events, state updates, subgoal switches, and the end of the trace receive $w_{t,i}=1$.
This weighting draws on SOM-DST~\cite{kim2020somdst}, which generates new values for dialogue-state slots that change between turns. Here, the VLM learns to generate the full trace, with greater weight on the changes.

\paragraph{Recoverable-history perturbations.}
During evaluation and GRPO rollouts, the write pass reads the VLM's own preceding state and subgoal.
Copycat~\cite{wen2020copycat} shows how access to observation histories can lead a policy to repeat its previous action. A similar shortcut is available when the write pass can copy its preceding state or subgoal.
In approximately 20\% of supervised training examples, we perturb one entry in the preceding state or replace the preceding subgoal, keeping the annotated target trace fixed.
A state perturbation shifts coordinates by approximately 10--30 pixels per axis, removes an entry, or changes an object's \texttt{held} status. A replacement subgoal comes from the same episode.
We perturb only values that the write-pass inputs can recover. For example, the identity of a cube hidden under a cup is preserved when the visual history no longer shows which cube was covered.

\subsection{Supervised Training}
\label{app:training}

We train supervised \rmr{} jointly on the trace and action losses for 10k steps with $\lambda=0.5$, then freeze the VLM and fine-tune the action expert for 5k steps. The peak learning rate is $2\times10^{-5}$.
\Cref{tab:trainingconfigs} specifies the seven schedules compared in \cref{tab:training}, including the five in Table~\ref{tab:training_main}.
Each schedule uses 15k optimization steps. Two begin with joint training, and five begin with 5k steps on the trace loss alone.

\begin{table*}[t]
    \centering
    \small
    \caption{\textbf{Training schedules.}
    All schedules use 15k optimization steps. The first two rows differ only in which model writes the trace at evaluation.
    Joint training followed by action-expert fine-tuning is supervised \rmr{}.}
    \label{tab:trainingconfigs}
    \begin{tabularx}{\textwidth}{@{}>{\raggedright\arraybackslash}p{.27\textwidth}Y@{}}
        \toprule
        Schedule & Training stages \\
        \midrule
        Sequential, two models & Trace loss, 5k $\rightarrow$ action loss, 10k, with the VLM updated. At evaluation, the model trained with the trace loss writes the trace and the model trained with the action loss acts. \\
        Sequential, single model & Same training as the two-model schedule. At evaluation, the model trained with the action loss both writes the trace and acts. \\
        Sequential, single model, frozen VLM & Trace loss, 5k $\rightarrow$ action loss on the action expert only, 10k, with the VLM frozen. \\
        Joint training throughout & Trace and action losses together, 15k. \\
        Trace loss, then joint training & Trace loss, 5k $\rightarrow$ trace and action losses together, 10k. \\
        Trace loss, action-expert training, then joint training & Trace loss, 5k $\rightarrow$ action loss on the action expert only, 5k, with the VLM frozen $\rightarrow$ trace and action losses together, 5k. \\
        Joint training, then action-expert fine-tuning & Trace and action losses together, 10k $\rightarrow$ action loss on the action expert only, 5k, with the VLM frozen. \\
        \bottomrule
    \end{tabularx}
\end{table*}

\paragraph{Trace-loss weight.}
Of the three trace-loss weights in \cref{tab:lambda}, $\lambda=0.5$ gives the highest overall success (59.71\%) and the highest scores on Counting, Reference, and Imitation. Permanence peaks at 67.00\% with $\lambda=1$.

\begin{table}[t]
    \centering
    \footnotesize
    \caption{\textbf{Trace-loss weight on RoboMME (success, \%).} All rows follow the supervised \rmr{} schedule and vary $\lambda$ during joint training. Bold marks column maxima.}
    \label{tab:lambda}
    \setlength{\tabcolsep}{2.6pt}
    \renewcommand{\arraystretch}{1.08}
    \begin{tabular}{@{}lrrrrr@{}}
        \toprule
        Trace-loss weight & Count. & Perm. & Ref. & Imit. & Avg. $\uparrow$ \\
        \midrule
        $\lambda=0.1$ & 63.83 & 51.17 & 72.17 & 41.67 & 57.21 \\
        $\lambda=0.5$ (\rmr{}) & \best{65.33} & 57.33 & \best{72.83} & \best{43.33} & \best{59.71} \\
        $\lambda=1$ & 53.50 & \best{67.00} & 63.67 & 31.67 & 53.96 \\
        \bottomrule
    \end{tabular}
\end{table}

\FloatBarrier
\section{GRPO Post-Training}
\label{app:posttraining}

Starting from supervised \rmr{}, we apply GRPO~\cite{shao2024deepseekmath,guo2025deepseekr1} to the generated traces. We train rank-32 LoRA adapters~\cite{hu2022lora} in the VLM at a learning rate of $3\times10^{-4}$, keeping the supervised base weights, visual encoder, and action expert frozen.
For each task layout, we sample $K=8$ episodes. An episode's advantage is its reward minus the mean reward of the other $K-1$ episodes (Eq.~\eqref{eq:advantage})~\cite{kool2019buy,ahmadian2024back}. A KL constraint of 0.1 regularizes the policy updates.

\paragraph{Episode reward.}
The episode reward combines the terminal outcome and subgoal correctness as $R=r_{\mathrm{term}}+\beta m$ (Eq.~\eqref{eq:reward}). The terminal reward $r_{\mathrm{term}}$ is $+1$ for success and $-1$ for failure or timeout.
Each act pass contributes one subgoal decision, and $m$ is the fraction of decisions in the episode that match their labels.
For the text comparison, we remove coordinates and the completion condition from both the generated subgoal and the label, then compare the remaining text in lowercase. When both specify a location, the coordinates must also agree within 8 pixels.
\Cref{tab:beta} extends Table~\ref{tab:rl} to three nonzero values of $\beta$. All three improve overall RoboMME success over GRPO with terminal reward alone. The setting used by \rmr{} + GRPO, $\beta=0.5$, gives the highest success in all four categories.

\begin{table}[t]
    \centering
    \footnotesize
    \caption{\textbf{Subgoal correctness weight in the episode reward on RoboMME (success, \%).} $\beta$ weights $m$ in $R=r_{\mathrm{term}}+\beta m$. GRPO trains the VLM's LoRA adapters, and the $\pi$RL-style baseline trains the action expert. Bold marks column maxima.}
    \label{tab:beta}
    \setlength{\tabcolsep}{2.6pt}
    \renewcommand{\arraystretch}{1.08}
    \begin{tabular}{@{}lrrrrr@{}}
        \toprule
        Configuration & Count. & Perm. & Ref. & Imit. & Avg. $\uparrow$ \\
        \midrule
        \rmr{} (supervised) & 65.33 & 57.33 & 72.83 & 43.33 & 59.71 \\
        \midrule
        $\pi$RL-style, terminal reward & 57.33 & 43.67 & 72.83 & 43.67 & 54.38 \\
        \midrule
        \multicolumn{6}{@{}l}{\textit{GRPO}} \\
        $\beta=0$ (terminal reward) & 65.33 & 57.83 & 76.17 & 46.17 & 61.38 \\
        $\beta=0.1$ & 67.17 & 59.33 & 77.17 & 46.17 & 62.46 \\
        $\beta=0.5$ (\rmr{} + GRPO) & \best{67.67} & \best{67.33} & \best{79.83} & \best{51.00} & \best{66.46} \\
        $\beta=1$ & 66.83 & 62.17 & 78.83 & 49.83 & 64.42 \\
        \bottomrule
    \end{tabular}
\end{table}

\FloatBarrier
\section{Benchmarks and Complete Results}
\label{app:results}

\subsection{Tasks}
\label{app:tasks}

\Cref{tab:robommetasks} describes the 16 RoboMME tasks and gives their abbreviations. \Cref{tab:liberomemtasks} describes the ten LIBERO-Mem tasks and gives the official step limit for each~\cite{liberomem}.

\begin{table*}[t]
    \centering
    \footnotesize
    \caption{\textbf{RoboMME tasks.}
    Task names and definitions follow RoboMME~\cite{robomme}.}
    \label{tab:robommetasks}
    \setlength{\tabcolsep}{4pt}
    \renewcommand{\arraystretch}{1.08}
    \begin{tabularx}{\textwidth}{@{}llY@{}}
        \toprule
        Task & Abbr. & Definition \\
        \midrule
        \multicolumn{3}{@{}l}{\textit{Counting}} \\
        BinFill & BF & Put the specified number of cubes of each color into the bin, then press the button. Cubes may appear over time. \\
        PickXtimes & PX & Pick up a cube of a given color and place it on the target a specified number of times, then press the button. \\
        SwingXtimes & SX & Carry a cube of a given color back and forth between the right- and left-side targets a specified number of times, put it down, then press the button. \\
        StopCube & SC & Press the button when a cube moving back and forth reaches the target on a specified visit. \\
        \midrule
        \multicolumn{3}{@{}l}{\textit{Permanence}} \\
        VideoUnmask & VU & Watch cups cover cubes in a video. Lift the cup hiding the specified cube, or the cups hiding two specified cubes in order. \\
        ButtonUnmask & BU & Press the button as cups cover the cubes. Lift the cup hiding the specified cube, or the cups hiding two specified cubes in order. \\
        VideoUnmaskSwap & VUS & Watch cups cover cubes and then swap positions. Lift the cup hiding the specified cube, or the cups hiding two specified cubes in order. \\
        ButtonUnmaskSwap & BUS & Press two buttons while cups cover the cubes and swap positions. Lift the cup hiding the specified cube, or the cups hiding two specified cubes in order. \\
        \midrule
        \multicolumn{3}{@{}l}{\textit{Reference}} \\
        PickHighlight & PH & Press the button and observe which cubes are briefly highlighted. Pick up each highlighted cube, then press the button again. \\
        VideoRepick & VR & Watch a cube being picked up and placed in a video. The cubes may then swap positions. Pick up the same cube the specified number of times, then press the button. \\
        VideoPlaceButton & VPB & Watch a video of cube placements interleaved with button presses. Place the cube at the target specified relative to a press, such as the last target before it. \\
        VideoPlaceOrder & VPO & Watch a cube being placed on a sequence of targets. Place it on the target at the specified position in the sequence, such as the third. \\
        \midrule
        \multicolumn{3}{@{}l}{\textit{Imitation}} \\
        MoveCube & MC & Move the cube to the target using the demonstrated technique (pick-and-place, pushing with the gripper, or hooking with a tool). \\
        InsertPeg & IP & Grasp the peg shown in a video at the same end and insert it into the box from the same side. \\
        PatternLock & PL & Use a stick to retrace the sequence of grid targets shown in a video. \\
        RouteStick & RS & Guide a stick around obstacles along the demonstrated path, circling each obstacle in the direction shown. \\
        \bottomrule
    \end{tabularx}
\end{table*}

\begin{table}[t]
    \centering
    \footnotesize
    \caption{\textbf{LIBERO-Mem tasks.}
    Each step limit is the maximum number of actions per episode under the official protocol~\cite{liberomem}.}
    \label{tab:liberomemtasks}
    \setlength{\tabcolsep}{3pt}
    \renewcommand{\arraystretch}{1.08}
    \begin{tabularx}{\columnwidth}{@{}llYr@{}}
        \toprule
        Task & Group & Definition & Steps \\
        \midrule
        T1 & Motion & Pick up the bowl and place it back on its plate. & 200 \\
        T2 & Motion & Lift the bottle and put it back on its plate. & 290 \\
        T3 & Sequence & Return the bowl to its plate three times. & 320 \\
        T4 & Sequence & Return the bottle to its plate three times. & 370 \\
        T5 & Sequence & Return the bowl to its plate five times. & 450 \\
        T6 & Sequence & Return the bowl to its plate seven times. & 655 \\
        T7 & Relation & Swap the two bowls between their plates using the empty plate. & 480 \\
        T8 & Relation & Rotate the three bowls across their plates from left to right using the empty plate. & 675 \\
        T9 & Occlusion & Put the cream cheese in the nearer basket, then move that basket to the center. & 390 \\
        T10 & Occlusion & Put the cream cheese in the nearer basket, then move the empty basket to the center. & 390 \\
        \bottomrule
    \end{tabularx}
\end{table}

\begin{table*}[t]
    \centering
    \caption{\textbf{RoboMME success by training schedule (\%).} All seven schedules use 15k supervised steps (\cref{tab:trainingconfigs}). Joint training followed by action-expert fine-tuning gives the supervised \rmr{} model in Table~\ref{tab:main}. Bold marks column maxima.}
    \label{tab:training}
    \scriptsize
    \setlength{\tabcolsep}{1.2pt}
    \renewcommand{\arraystretch}{1.12}
    \begin{tabular*}{\textwidth}{@{\extracolsep{\fill}}>{\raggedright\arraybackslash}p{.24\textwidth}*{17}{r}@{}}
        \toprule
        Schedule & \multicolumn{4}{c}{Counting} & \multicolumn{4}{c}{Permanence} & \multicolumn{4}{c}{Reference} & \multicolumn{4}{c}{Imitation} & Avg. $\uparrow$ \\
        \cmidrule(lr){2-5}\cmidrule(lr){6-9}\cmidrule(lr){10-13}\cmidrule(lr){14-17}
         & BF & PX & SX & SC & VU & BU & VUS & BUS & PH & VR & VPB & VPO & MC & IP & PL & RS & \\
        \midrule
        Sequential, two models & 52.67 & 54.67 & 25.33 & 16.00 & \best{80.67} & 60.67 & 24.67 & 27.33 & 67.33 & 42.67 & 84.67 & 84.00 & \best{82.67} & 6.00 & 30.00 & 13.33 & 47.04 \\
        Sequential, single model & 4.00 & 7.33 & 0.00 & 10.67 & 20.67 & 0.00 & 16.00 & 0.00 & 0.00 & 0.00 & 16.00 & 21.33 & 12.00 & 0.00 & 20.67 & 4.00 & 8.29 \\
        Sequential, single model, frozen VLM & 9.33 & 29.33 & 4.00 & 14.00 & 45.33 & 62.00 & 22.00 & 21.33 & 30.00 & 14.00 & 67.33 & 59.33 & 52.00 & 0.00 & 10.00 & 0.00 & 27.50 \\
        Joint training throughout & 62.67 & 86.67 & 68.00 & \best{16.67} & 74.00 & 70.67 & 24.67 & 32.00 & \best{69.33} & 46.00 & 84.67 & \best{87.33} & 76.67 & 4.00 & 31.33 & 43.33 & 54.88 \\
        Trace loss, then joint training & 58.67 & 96.00 & 54.67 & 10.67 & 60.00 & 58.67 & 21.33 & 23.33 & \best{69.33} & 48.67 & 85.33 & \best{87.33} & 73.33 & 3.33 & \best{36.67} & 36.00 & 51.46 \\
        Trace loss, action-expert training, then joint training & 46.00 & 78.00 & 46.00 & 4.00 & 72.00 & 76.00 & 21.33 & 24.67 & 59.33 & 44.00 & 81.33 & 78.67 & 63.33 & 4.67 & 20.67 & 28.00 & 46.75 \\
        Joint training, then action-expert fine-tuning & \best{66.67} & \best{100.00} & \best{82.00} & 12.67 & 78.67 & \best{81.33} & \best{31.33} & \best{38.00} & 63.33 & \best{54.00} & \best{86.67} & \best{87.33} & \best{82.67} & \best{8.67} & 34.67 & \best{47.33} & \best{59.71} \\
        \bottomrule
    \end{tabular*}
\end{table*}

\begin{table*}[t]
    \centering
    \caption{\textbf{RoboMME success by trace variant (\%).} The first row writes all four fields without recurrent input. The next four rows start from the grounded subgoal $g_t$ and add fields cumulatively, as in Table~\ref{tab:structure}. The last four rows specify their fields explicitly. In all recurrent variants, the subgoal and any maintained state carry over to the next update. Bold marks column maxima.}
    \label{tab:structurefull}
    \scriptsize
    \setlength{\tabcolsep}{1.2pt}
    \renewcommand{\arraystretch}{1.12}
    \begin{tabular*}{\textwidth}{@{\extracolsep{\fill}}>{\raggedright\arraybackslash}p{.24\textwidth}*{17}{r}@{}}
        \toprule
        Trace & \multicolumn{4}{c}{Counting} & \multicolumn{4}{c}{Permanence} & \multicolumn{4}{c}{Reference} & \multicolumn{4}{c}{Imitation} & Avg. $\uparrow$ \\
        \cmidrule(lr){2-5}\cmidrule(lr){6-9}\cmidrule(lr){10-13}\cmidrule(lr){14-17}
         & BF & PX & SX & SC & VU & BU & VUS & BUS & PH & VR & VPB & VPO & MC & IP & PL & RS & \\
        \midrule
        Without recurrence & 36.67 & 44.67 & 33.33 & 6.00 & 58.00 & 50.67 & 17.33 & 14.67 & 35.33 & 22.00 & 32.67 & 41.33 & 59.33 & 2.67 & 8.00 & 13.33 & 29.75 \\
        \midrule
        Subgoal $g_t$ only & 26.00 & 34.00 & 51.33 & 6.67 & 62.67 & 67.33 & 17.33 & 22.67 & 66.67 & 18.67 & 20.00 & 23.33 & 64.00 & 0.67 & 11.33 & 20.67 & 32.08 \\
        + state $s_t$ & 54.00 & 93.33 & 64.00 & 8.00 & 74.67 & 63.33 & 16.67 & 23.33 & 54.00 & 42.00 & 73.33 & 65.33 & 77.33 & 4.67 & 26.00 & 34.67 & 48.42 \\
        + evidence $e_t$ & \best{66.67} & 96.67 & 78.67 & 8.67 & 76.00 & 80.00 & 21.33 & 31.33 & 58.67 & 45.33 & 80.67 & 84.00 & 78.00 & 6.00 & 32.67 & \best{47.33} & 55.75 \\
        + event $v_t$ (\rmr{}) & \best{66.67} & \best{100.00} & \best{82.00} & \best{12.67} & 78.67 & 81.33 & 31.33 & \best{38.00} & 63.33 & 54.00 & \best{86.67} & 87.33 & \best{82.67} & \best{8.67} & \best{34.67} & \best{47.33} & \best{59.71} \\
        \midrule
        $v_t$, $s_t$, $g_t$ & 60.67 & 99.33 & 48.67 & 11.33 & \best{82.00} & \best{88.67} & \best{38.67} & 36.00 & 62.67 & \best{62.00} & 80.67 & \best{88.00} & 78.00 & 2.00 & 31.33 & 46.00 & 57.25 \\
        $v_t$, $g_t$ & 27.33 & 36.00 & 72.67 & 5.33 & 71.33 & 78.00 & 20.00 & 21.33 & \best{68.67} & 27.33 & 32.67 & 14.00 & 62.67 & 0.00 & 20.67 & 18.00 & 36.00 \\
        $e_t$, $v_t$, $g_t$ & 22.67 & 32.00 & 41.33 & 4.00 & 80.00 & 76.67 & 32.00 & 24.00 & 59.33 & 15.33 & 22.00 & 19.33 & 68.67 & 0.00 & 17.33 & 18.00 & 33.29 \\
        $e_t$, $g_t$ & 20.67 & 36.00 & 38.00 & 4.00 & 76.00 & 79.33 & 28.67 & 26.00 & 60.67 & 16.00 & 24.00 & 17.33 & 62.00 & 4.00 & 18.67 & 20.00 & 33.21 \\
        \bottomrule
    \end{tabular*}
\end{table*}

\subsection{Evaluation and Subgoal Diagnostics}
\label{app:evaluation}

For each of three evaluation seeds, we run 50 episodes per RoboMME task and 40 per LIBERO-Mem task, following the official protocols.
A task's success rate is the fraction of successful episodes pooled across the three seeds. Each RoboMME category averages its four task scores. Overall success gives equal weight to every task in the benchmark.
We compute differences before rounding. The supervised \rmr{} model is the common reference for the trace, training schedule, update interval, and trace-loss weight comparisons.

\paragraph{Subgoal diagnostics.}
Table~\ref{tab:structure} measures whether each generated subgoal specifies the decision required by its task category.
\begin{itemize}[leftmargin=*,nosep]
    \item \textbf{Count/stop (Counting).} At each decision, we compare the subgoal and label using text alone to check the repetition index and when to stop.
    \item \textbf{Container (Permanence).} We score the first cup selection in each episode. A selection is correct when both the named target identity and the cup location match the label.
    \item \textbf{Cue target (Reference).} We score decisions that select a highlighted or demonstrated cube, or place a cube at a demonstrated target. Both the subgoal text and the target location must match the label. Only episodes that reach an eligible decision contribute.
    \item \textbf{Demo step (Imitation).} At each decision, we compare the subgoal and label using text alone to check whether the action follows the demonstrated sequence, including direction, route, and tool use.
\end{itemize}

\paragraph{Overall subgoal correctness.}
For each episode, $m$ is the fraction of all subgoal decisions that match their labels under the GRPO reward criterion (\cref{app:posttraining}). BinFill pickup decisions use text alone because any cube of the required color is a valid target.
We average $m$ over episodes within each task and give the 16 RoboMME tasks equal weight in the overall score.

\begin{table*}[t]
    \centering
    \caption{\textbf{Task success for additional RoboMME configurations (\%).} Scores follow each method's published evaluation protocol. GroundSG and SimpleSG condition \pizero{} on grounded or simple language subgoals, respectively, supplied by the VLM named in each row~\cite{robomme}. The \rmr{} scores are from Table~\ref{tab:main}.}
    \label{tab:additional}
    \scriptsize
    \setlength{\tabcolsep}{1.2pt}
    \renewcommand{\arraystretch}{1.12}
    \begin{tabular*}{\textwidth}{@{\extracolsep{\fill}}>{\raggedright\arraybackslash}p{.24\textwidth}*{17}{r}@{}}
        \toprule
        Method & \multicolumn{4}{c}{Counting} & \multicolumn{4}{c}{Permanence} & \multicolumn{4}{c}{Reference} & \multicolumn{4}{c}{Imitation} & Avg. $\uparrow$ \\
        \cmidrule(lr){2-5}\cmidrule(lr){6-9}\cmidrule(lr){10-13}\cmidrule(lr){14-17}
         & BF & PX & SX & SC & VU & BU & VUS & BUS & PH & VR & VPB & VPO & MC & IP & PL & RS & \\
        \midrule
        \multicolumn{18}{@{}l}{\textit{With an external reasoning model}} \\
        GroundSG+QwenVL~\cite{robomme} & 52.00 & 92.67 & 7.33 & 0.00 & 88.67 & 24.00 & 30.67 & 14.00 & 15.11 & 25.33 & 54.00 & 31.78 & 71.56 & 3.33 & 6.67 & 6.00 & 32.70 \\
        GroundSG+Gemini-2.5-Pro~\cite{robomme} & 26.00 & 18.00 & 4.00 & 3.00 & 36.00 & 14.00 & 13.00 & 0.00 & 9.00 & 17.00 & 12.00 & 7.00 & 17.00 & 0.00 & 7.00 & 2.00 & 11.56 \\
        SimpleSG+Gemini-2.5-Pro~\cite{robomme} & 46.00 & 63.00 & 45.00 & 2.00 & 29.00 & 9.00 & 14.00 & 2.00 & 20.00 & 15.00 & 26.00 & 29.00 & 61.00 & 4.00 & 7.00 & 0.00 & 23.25 \\
        GPT-6-astra+VLA~\cite{pal} & 38.00 & 88.00 & 44.00 & 6.00 & 78.00 & 32.00 & 28.00 & 0.00 & 74.00 & 32.00 & 50.00 & 82.00 & 54.00 & 2.00 & 24.00 & 4.00 & 39.75 \\
        SERA~\cite{sera} & 48.00 & 70.00 & 72.00 & 6.00 & 96.00 & 40.00 & 92.00 & 0.00 & 54.00 & 34.00 & 98.00 & 92.00 & 60.00 & 0.00 & 60.00 & 0.00 & 51.38 \\
        \midrule
        \multicolumn{18}{@{}l}{\textit{Without an external reasoning model}} \\
        D\&R (64 tokens)~\cite{divide} & 35.33 & 82.67 & 95.33 & 31.33 & 32.67 & 26.00 & 27.33 & 21.33 & 24.00 & 26.67 & 34.00 & 29.33 & 56.67 & 3.33 & 32.00 & 59.33 & 38.58 \\
        \textbf{\rmr{} (supervised)} & 66.67 & 100.00 & 82.00 & 12.67 & 78.67 & 81.33 & 31.33 & 38.00 & 63.33 & 54.00 & 86.67 & 87.33 & 82.67 & 8.67 & 34.67 & 47.33 & 59.71 \\
        \textbf{\rmr{} + GRPO} & 73.33 & 99.33 & 92.00 & 6.00 & 94.67 & 93.33 & 30.67 & 50.67 & 69.33 & 72.67 & 86.67 & 90.67 & 90.67 & 5.33 & 56.67 & 51.33 & 66.46 \\
        \bottomrule
    \end{tabular*}
\end{table*}

\subsection{Training Schedules}
\label{app:schedules}

With 15k optimization steps for each schedule (\cref{tab:trainingconfigs,tab:training}), supervised \rmr{} reaches 59.71\% success, 4.83 points above joint training throughout (54.88\%).
Ending joint training with action-expert fine-tuning gives the highest average in all four categories and the highest or joint-highest score on 12 of 16 tasks.
Starting with 5k steps of trace supervision and then training jointly for 10k steps reaches 51.46\%. Splitting those final 10k steps equally between action-expert training and joint training, in that order, reaches 46.75\%.

\subsection{Trace Ablation}
\label{app:traceablation}

\Cref{tab:structurefull} reports success on all 16 RoboMME tasks for each trace variant. \Cref{fig:structure_radar} plots the five variants from Table~\ref{tab:structure}.
The variant \emph{Without recurrence} writes all four fields and receives no trace from the preceding update. Each of the eight recurrent variants receives the preceding grounded subgoal and, if it writes a maintained state, the preceding state as well.

The four recurrent variants that carry the maintained state reach 48.42--59.71\% overall success. The four that carry only the subgoal reach 32.08--36.00\%, across the same choices of observation evidence and typed event.
Starting from state and subgoal alone (48.42\%), adding observation evidence raises success to 55.75\%, and adding the typed event alone raises it to 57.25\%. The full four-field trace reaches 59.71\%.

\input{content/figures/rmr_trace_ablation_radar}
\begin{figure*}[t]
    \centering
    \footnotesize
    \setlength{\tabcolsep}{2.5pt}
    \renewcommand{\arraystretch}{1.0}
    \begin{tabularx}{\textwidth}{@{}>{\raggedright\arraybackslash}p{0.028\textwidth}*{5}{Y}@{}}
        \toprule
        & \multicolumn{1}{c}{$t=9$} & \multicolumn{1}{c}{$t=10$} & \multicolumn{1}{c}{$t=11$} & \multicolumn{1}{c}{$t=14$} & \multicolumn{1}{c}{$t=15$} \\
        \midrule
        $o_t$
            & \exframe{\includegraphics[width=\linewidth,trim={20bp 106bp 20bp 10bp},clip]{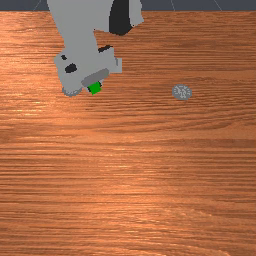}}{20}{106}{20}{10}{216}{\exring{rmrgoal}{86}{96}}
            & \exframe{\includegraphics[width=\linewidth,trim={20bp 106bp 20bp 10bp},clip]{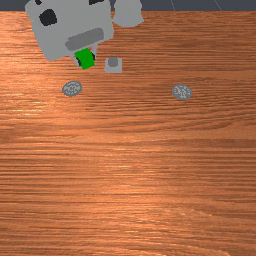}}{20}{106}{20}{10}{216}{\exring{rmrgoal}{90}{75}}
            & \exframe{\includegraphics[width=\linewidth,trim={20bp 106bp 20bp 10bp},clip]{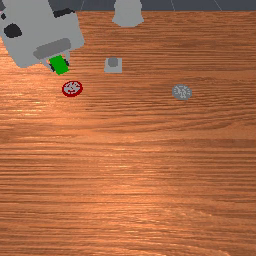}}{20}{106}{20}{10}{216}{\exring{rmrgoal}{94}{189}}
            & \exframe{\includegraphics[width=\linewidth,trim={20bp 106bp 20bp 10bp},clip]{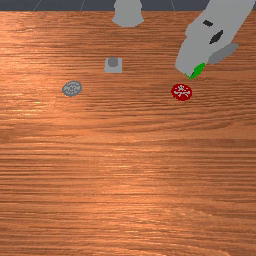}}{20}{106}{20}{10}{216}{}
            & \exframe{\includegraphics[width=\linewidth,trim={20bp 106bp 20bp 10bp},clip]{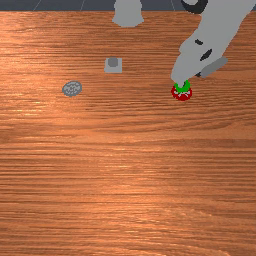}}{20}{106}{20}{10}{216}{\exring{rmrgoal}{62}{113}} \\
        \addlinespace[2pt]
        \textcolor{rmrevidence}{$\boldsymbol{e_t}$}
            & \tracecell{grasped green cube}
            & \tracecell{done: picked up green cube}
            & \tracecell{done: reached right target}
            & \tracecell{done: reached left target}
            & \tracecell{done: put down; released green cube} \\
        \textcolor{rmrevent}{$\boldsymbol{v_t}$}
            & \tracecell{none} & \tracecell{none} & \tracecell{none} & \tracecell{none} & \tracecell{none} \\
        \textcolor{rmrstate}{$\boldsymbol{s_t}$}
            & \tracecell{step 0; button <62,113>; green cube <86,96> held; right target <90,75>; left target <94,189>; swung 0 of 2}
            & \tracecell{step 1; button <62,113>; right target <90,75>; left target <94,189>; swung 0 of 2}
            & \tracecell{step 2; button <62,113>; right target <90,75>; left target <94,189>; swung 1 of 2}
            & \tracecell{step 3; button <62,113>; green cube <86,96> held; right target <90,75>; left target <94,189>; swung 2 of 2}
            & \tracecell{step 4; button <62,113>; right target <90,75>; left target <94,189>; swung 2 of 2} \\
        \textcolor{rmrgoal}{$\boldsymbol{g_t}$}
            & \tracecell{pick up the green cube at <86,96> until picked up}
            & \tracecell{move to the top of the right-side target at <90,75> for the first time until reached}
            & \tracecell{move to the top of the left-side target at <94,189> for the first time until reached}
            & \tracecell{put the green cube on the table until put down}
            & \tracecell{press the button at <62,113> until pressed} \\
        \bottomrule
    \end{tabularx}
    \caption{\textbf{The swing count determines when to put the cube down.}
    In this successful RoboMME SwingXtimes episode, supervised \rmr{} records each arrival in the state.
    After one swing the subgoal directs the robot to the opposite target. After two, it directs the robot to put the cube down.
    Green rings mark subgoal locations.}
    \label{fig:example_swing}
\end{figure*}

\subsection{Published Results}
\label{app:published}

\paragraph{RoboMME.}
For MemoryVLA, the overall score in Table~\ref{tab:main} is the mean of the 16 task scores reported by the RoboMME authors~\cite{robomme}.
\Cref{tab:additional} gives per-task results for the 64-token D\&R model and additional systems that predict subgoals.

\paragraph{LIBERO-Mem.}
Table~\ref{tab:liberomem} gives published overall scores and, where available, task scores from the same evaluation.
For the two entries marked $^{\dagger}$, relaxed success credits completion of the full ordered goal even if the policy repeats an action afterward~\cite{twoam}.

\section{Qualitative Examples}
\label{app:examples}

\begin{figure*}[t]
    \centering
    \footnotesize
    \setlength{\tabcolsep}{2.5pt}
    \renewcommand{\arraystretch}{1.0}
    \begin{tabularx}{\textwidth}{@{}>{\raggedright\arraybackslash}p{0.028\textwidth}*{5}{Y}@{}}
        \toprule
        & \multicolumn{3}{c}{\textit{Demonstration video}} & \multicolumn{2}{c}{\textit{Execution}} \\
        \cmidrule(lr){2-4}\cmidrule(l){5-6}
        & \multicolumn{1}{c}{$t=7$} & \multicolumn{1}{c}{$t=11$} & \multicolumn{1}{c}{$t=16$} & \multicolumn{1}{c}{$t=19$} & \multicolumn{1}{c}{$t=30$} \\
        \midrule
        $o_t$
            & \exframe{\includegraphics[width=\linewidth,trim={28bp 106bp 28bp 20bp},clip]{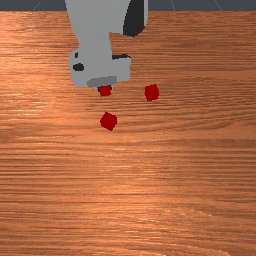}}{28}{106}{28}{20}{200}{\exring{rmrstate}{90}{106}}
            & \exframe{\includegraphics[width=\linewidth,trim={28bp 106bp 28bp 20bp},clip]{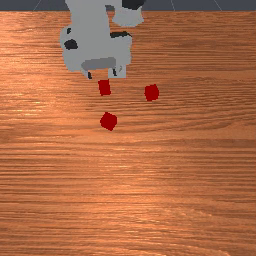}}{28}{106}{28}{20}{200}{\exring{rmrstate}{86}{103}}
            & \exframe{\includegraphics[width=\linewidth,trim={28bp 106bp 28bp 20bp},clip]{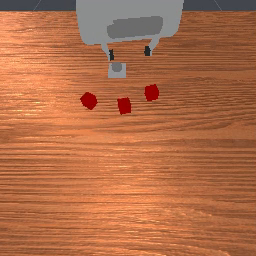}}{28}{106}{28}{20}{200}{\exring{rmrstate}{103}{121}}
            & \exframe{\includegraphics[width=\linewidth,trim={28bp 106bp 28bp 20bp},clip]{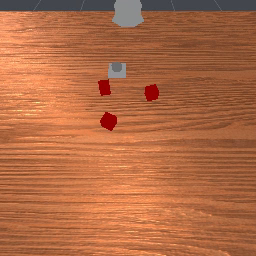}}{28}{106}{28}{20}{200}{\exring{rmrgoal}{120}{109}}
            & \exframe{\includegraphics[width=\linewidth,trim={28bp 106bp 28bp 20bp},clip]{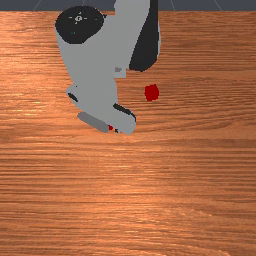}}{28}{106}{28}{20}{200}{} \\
        \addlinespace[2pt]
        \textcolor{rmrevidence}{$\boldsymbol{e_t}$}
            & \tracecell{grasped cube\_A}
            & \tracecell{shown: drop the cube on the table}
            & \tracecell{moving cube\_A, cube\_B}
            & \tracecell{moving cube\_A, cube\_B}
            & \tracecell{done: put down; released cube\_A} \\
        \textcolor{rmrevent}{$\boldsymbol{v_t}$}
            & \tracecell{none} & \tracecell{none} & \tracecell{none}
            & \tracecell{swap\#1 cube\_A-cube\_B}
            & \tracecell{none} \\
        \textcolor{rmrstate}{$\boldsymbol{s_t}$}
            & \tracecell{step 0; button <66,117>; cube\_A <90,106> held; target cube\_A; repicked 0 of 1}
            & \tracecell{step 0; button <66,117>; cube\_A <86,103>; target cube\_A; repicked 0 of 1}
            & \tracecell{step 0; button <66,117>; cube\_A <103,121> moving; target cube\_A; repicked 0 of 1}
            & \tracecell{step 0; button <66,117>; cube\_A <116,108>; target cube\_A; repicked 0 of 1}
            & \tracecell{step 2; button <66,117>; cube\_A <130,109>; target cube\_A; repicked 1 of 1} \\
        \textcolor{rmrgoal}{$\boldsymbol{g_t}$}
            & \tracecell{watch}
            & \tracecell{watch}
            & \tracecell{watch}
            & \tracecell{pick up the correct cube at <120,109> for the first time until picked up}
            & \tracecell{press the button to finish until pressed} \\
        \bottomrule
    \end{tabularx}
    \caption{\textbf{The target identity persists through a swap.}
    In this successful RoboMME VideoRepick episode, supervised \rmr{} identifies the demonstrated cube as \texttt{cube\_A} among three identical cubes.
    Purple rings mark its location in the state during the demonstration. The green ring marks the pickup location specified by the first execution subgoal.}
    \label{fig:example_repick}
\end{figure*}

\begin{figure*}[t]
    \centering
    \footnotesize
    \setlength{\tabcolsep}{2.5pt}
    \renewcommand{\arraystretch}{1.0}
    \begin{tabularx}{\textwidth}{@{}>{\raggedright\arraybackslash}p{0.028\textwidth}*{5}{Y}@{}}
        \toprule
        & \multicolumn{1}{c}{$t=9$} & \multicolumn{1}{c}{$t=13$} & \multicolumn{1}{c}{$t=14$} & \multicolumn{1}{c}{$t=19$} & \multicolumn{1}{c}{$t=21$} \\
        \midrule
        $o_t$
            & \exframe{\includegraphics[width=\linewidth,trim={16bp 71bp 16bp 40bp},clip]{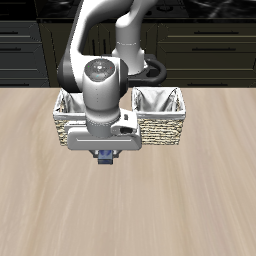}}{16}{71}{16}{40}{224}{\exring{rmrgoal}{134}{85}}
            & \exframe{\includegraphics[width=\linewidth,trim={16bp 71bp 16bp 40bp},clip]{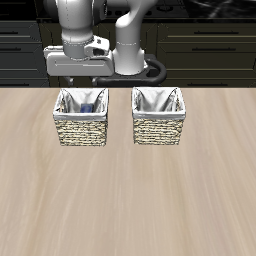}}{16}{71}{16}{40}{224}{\exring{rmrgoal}{134}{85}}
            & \exframe{\includegraphics[width=\linewidth,trim={16bp 71bp 16bp 40bp},clip]{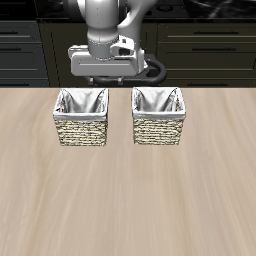}}{16}{71}{16}{40}{224}{\exring{rmrgoal}{134}{85}}
            & \exframe{\includegraphics[width=\linewidth,trim={16bp 71bp 16bp 40bp},clip]{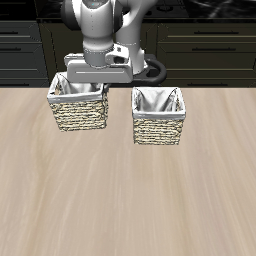}}{16}{71}{16}{40}{224}{\exring{rmrgoal}{160}{102}}
            & \exframe{\includegraphics[width=\linewidth,trim={16bp 71bp 16bp 40bp},clip]{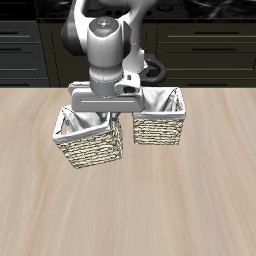}}{16}{71}{16}{40}{224}{} \\
        \addlinespace[2pt]
        \textcolor{rmrevidence}{$\boldsymbol{e_t}$}
            & \tracecell{done: picked up the cream cheese}
            & \tracecell{still}
            & \tracecell{done: put the cream cheese in basket 1}
            & \tracecell{done: picked up basket 1}
            & \tracecell{done: moved basket 1 to the center} \\
        \textcolor{rmrevent}{$\boldsymbol{v_t}$}
            & \tracecell{none}
            & \tracecell{cover\#1 basket 1 covers the cream cheese}
            & \tracecell{none} & \tracecell{none} & \tracecell{none} \\
        \textcolor{rmrstate}{$\boldsymbol{s_t}$}
            & \tracecell{step 1; the cream cheese <147,106> held; basket 1 <134,85>; basket 2 <133,153>; placed 0 of 2}
            & \tracecell{step 1; the cream cheese in basket 1; basket 1 <134,85>; basket 2 <133,153>; placed 0 of 2}
            & \tracecell{step 2; the cream cheese in basket 1; basket 1 <134,85>; basket 2 <133,153>; placed 1 of 2}
            & \tracecell{step 3; the cream cheese in basket 1; basket 1 <118,85> held; basket 2 <133,153>; placed 1 of 2}
            & \tracecell{step 4; the cream cheese in basket 1; basket 1 <152,101> moving; basket 2 <133,153>; placed 2 of 2} \\
        \textcolor{rmrgoal}{$\boldsymbol{g_t}$}
            & \tracecell{put the cream cheese in basket 1 at <134,85> until placed}
            & \tracecell{put the cream cheese in basket 1 at <134,85> until placed}
            & \tracecell{pick up basket 1 at <134,85> until picked up}
            & \tracecell{move basket 1 to the center at <160,102> until placed}
            & \tracecell{done} \\
        \bottomrule
    \end{tabularx}
    \caption{\textbf{The state records which basket holds the cream cheese.}
    In this successful LIBERO-Mem T9 episode, supervised \rmr{} keeps the association with basket~1 after the cheese is hidden.
    The next two subgoals direct the robot to pick up that basket and move it to the center.
    Green rings mark subgoal locations.}
    \label{fig:example_basket}
\end{figure*}

\paragraph{Tracking a hidden cube in Figure~\ref{fig:teaser}.}
In ButtonUnmaskSwap, the state associates each hidden cube with the image coordinates of its cup.
Between \emph{Maintain} and \emph{Update}, the first dashed link spans four intervening updates. The cup hiding green trades places with the empty cup during this interval, changing green's location in the state.
At \emph{Update}, the event records the swap of the cups hiding red and blue while the subgoal is still to press the left button.
The second dashed link spans three intervening updates. At \emph{Ground}, four updates after the swap, completion of the button press switches the subgoal to lifting the cup at red's current location (Table~\ref{tab:trace}).
The \emph{Execute} column shows the robot revealing red after repeated action chunks and memory updates.
The G/R/B overlays mark cube identities in the camera views. The figure abbreviates the model's evidence, event, and subgoal.

\paragraph{Reading the traces.}
\Cref{fig:example_swing,fig:example_repick,fig:example_basket} show three successful evaluation episodes of supervised \rmr{}.
Each column pairs the front-camera view with the four fields generated at that update, quoted verbatim. Update indices $t$ are counted from the start of the episode.
The views are cropped for display. As in Figure~\ref{fig:teaser}, coordinates $\langle y,x\rangle$ index rows and columns of the original $256\!\times\!256$ image.

\paragraph{Counting.}
In SwingXtimes (\cref{fig:example_swing}), the step counter tracks completed subgoals and the swing count tracks arrivals at the two targets.
At $t=9$, contact evidence marks the cube as held. At $t=10$, pickup completion advances the step counter and the next subgoal directs the robot to the right-side target.
Reaching that target ($t=11$) advances the count to \texttt{swung 1 of 2} and switches the subgoal to the left-side target. Reaching the left target ($t=14$) completes the second swing, so the next subgoal is to put the cube down.
Once the cube is released ($t=15$), the subgoal switches to pressing the button. The event remains \texttt{none} at all five updates.

\paragraph{Reference.}
In VideoRepick (\cref{fig:example_repick}), the demonstration singles out one of three identical cubes.
While the subgoal is \texttt{watch}, the state identifies the grasped cube as the target, \texttt{cube\_A} ($t=7$). The evidence records the demonstrated drop ($t=11$), and the state updates the target's location as two cubes move ($t=16$).
At the first execution update ($t=19$), the event names the swap and the subgoal directs the action expert to pick up the demonstrated cube at its new location.
After the robot puts the cube down ($t=30$), the state records \texttt{repicked 1 of 1} and the next subgoal is to press the button.

\paragraph{Occlusion.}
LIBERO-Mem T9 (\cref{fig:example_basket}) requires putting the cream cheese into the nearer of two identical baskets and moving that basket to the center.
The subgoal selects basket~1 at $t=9$. When the cheese disappears inside it ($t=13$), the event records the occlusion and the state replaces the cheese's coordinates with \texttt{the cream cheese in basket 1}.
Completion evidence advances the placement count to \texttt{placed 1 of 2} at $t=14$, and the subgoal switches to picking up basket~1.
The state retains the association as the robot moves the basket to the center. At $t=21$, the count reaches \texttt{placed 2 of 2} and the subgoal is \texttt{done}.

\end{document}